\documentclass[letterpaper]{article} 
\usepackage{aaai2026}
\usepackage{times}  
\usepackage{helvet}  
\usepackage{courier}  
\usepackage[hyphens]{url}  
\usepackage{graphicx} 
\usepackage{natbib}  
\usepackage{caption} 
\usepackage{algorithm}
\usepackage[algo2e]{algorithm2e}
\usepackage{booktabs}
\usepackage{tikz}
\usepackage{mathrsfs}

\usepackage{newfloat}
\usepackage{listings}
\DeclareCaptionStyle{ruled}{labelfont=normalfont,labelsep=colon,strut=off} 
\floatstyle{ruled}
\newfloat{listing}{tb}{lst}{}
\floatname{listing}{Listing}
\nocopyright

\title{Breaking the Dyadic Barrier: Rethinking Fairness in Link Prediction Beyond Demographic Parity}\author{
    João Mattos$^*$,
    Debolina Halder Lina$^*$,
    Arlei Silva$^{*,\dagger}$
}
\affiliations{
    $^*$Computer Science Department, Rice University; $^\dagger$Ken Kennedy Institute, Rice University

    Houston, TX, USA\\
    \{jrm28,dl73,arlei\}@rice.edu
}

\usepackage{multirow}
\usepackage{multicol}
\usepackage{xcolor}
\definecolor{blueish}{HTML}{157fd4}
\definecolor{redish}{HTML}{d42015}
\definecolor{purpleish}{HTML}{755075}

\usepackage{amsmath}
\usepackage{amsthm}
\usepackage{cleveref}

\theoremstyle{definition}
\newtheorem{definition}{Definition}

\newtheorem{property}{Property}

\newtheorem{theorem}{Theorem}

\newcommand{\dempar}{$\Delta_{DP}$ }

\newcommand{\Vs}{s'}      
\newcommand{\Vns}{s}  

\newcommand{\ESNS}{E_{\Vs \text{-} \Vns}}  
\newcommand{\ESS}{E_{\Vs \text{-} \Vs}}    
\newcommand{\ENSNS}{E_{\Vns \text{-} \Vns}}  

\begin{document}

\maketitle

\begin{abstract}
Link prediction is a fundamental task in graph machine learning with applications ranging from social recommendation to knowledge graph completion. Fairness in this setting is critical, as biased predictions can exacerbate societal inequalities. Prior work adopts a dyadic definition of fairness, enforcing fairness through demographic parity between intra-group and inter-group link predictions. However, we show that this dyadic framing can obscure underlying disparities across subgroups, allowing systemic biases to go undetected. Moreover, we argue that demographic parity does not meet the desired properties for fairness assessment in ranking-based tasks such as link prediction. We formalize the limitations of existing fairness evaluations and propose a framework that enables a more expressive assessment. Additionally, we propose a lightweight post-processing method combined with decoupled link predictors that effectively mitigates bias and achieves state-of-the-art fairness–utility trade-offs.\footnote{This paper has been accepted for publication at the AAAI Conference on Artificial Intelligence (AAAI 2026).}
\end{abstract}

\begin{links}
    \link{Code}{https://github.com/joaopedromattos/MORAL}
\end{links}

\section{Introduction}



Link prediction is the task of discovering potential missing or future links in a network. There is extensive literature on link prediction models~\cite{liben2007link,zhang_link_2018,pan2021neural,zhu2021neural,chamberlain2022graph}, which demonstrates the relevance of this task and the variety of domains where it can be applied, ranging from social networks to knowledge graphs. Among these domains, many types of networks are prone to biases in their structure and features~\cite{dai2024comprehensive,stoica2018algorithmic,karimi2018homophily}, requiring the adoption of fair machine learning methods to mitigate bias in model predictions~\cite{li_dyadic_2021,li_fairlp_2022,current_fairegm_2022, tsioutsiouliklis2022link,tsioutsiouliklis2021fairness}. Our work is focused on the fair link prediction problem and, more specifically, on the evaluation and design of link prediction models under fairness considerations.

As a motivational example, consider the recommendation problem in social networks, where links represent social interactions. On a professional network, such as LinkedIn, biased connection recommendations can lead to persistent employment disparities between groups. Closed male networking circles provide more job leads and higher status connections than female/minority ones ~\cite{calvo2004effects}, resulting in white women receiving 33\% fewer job leads on average, according to some models \cite{mcdonald2009networks}. In addition, bias in friendship can lead to long-term accumulation \cite{gupta2021correcting,hofstra2017sources}, with the large majority of links occurring only within the same communities, creating filter bubbles \cite{cinelli2021echo,bakshy2015exposure}. 

The graph representation learning literature has focused on fair node representations~\cite{zhu_one_2024,zhu_devil_2024, ling_learning_2023, agarwal_towards_2021,laclau2021all}, and building upon these works, the current notion of fairness adopted by fair link prediction methods is \textit{dyadic}~\cite{li_dyadic_2021,current_fairegm_2022,li_fairlp_2022,luo2023cross}. The main assumption behind dyadic fairness is to categorize groups according to a protected attribute of interest and obtain equalized positive outcomes across these groups. In the social network example, two communities can be identified (e.g., men and women), and we can define fairness as links between men and women (inter) having the same probability of occurring as links within these groups (intra). To mitigate the disparity between these probabilities, fair link prediction algorithms adopt different strategies, evaluated through group fairness statistical metrics~\cite{dwork_fairness_2011,kusner_counterfactual_2017,masrour_bursting_2020}. 

However, we show that fair node embeddings do not translate to fair link prediction. The limited expressive power of Graph Neural Networks (GNNs), outputs indistinguishable node representations in symmetric neighborhoods containing different sensitive groups. This prevents the adoption of fair node embeddings in training objectives that distinguish between edge groups to achieve parity.

In addition, the dyadic fairness applied by previous fair link prediction methods (e.g., intra vs. inter) is unable to capture biases that occur \textit{within} sensitive groups of node pairs, which is known as fairness gerrymandering  \cite{kearns2018preventing}. In the social network example, male-male connections might be systematically more likely than female-female connections, and the dyadic fairness evaluation metrics considered (in
particular, demographic parity) are insensitive to this phenomenon, thereby masking underlying biases. The consequence of this limitation is a "glass ceiling" effect~\cite{stoica2018algorithmic}, in which an under-representation of a subgroup of pairs goes undetected.

A third limitation of existing work on fair graph machine learning (including link prediction) is evaluating fairness as a subset selection problem, instead of based on ranking~\cite{hanretiring,kleinberg2024calibrated,stoica2024fairness,li2021user}. We claim that by not considering the ranking of candidate links, a link prediction method can be prone to \textit{exposure bias}. In this fashion, one key property of a fair link prediction algorithm should be to ensure equality of probabilities and exposure, preventing one group from dominating the top positions of the ranking.

Our work is the first attempt at formalizing and demonstrating empirically the above limitations in the context of fair link prediction. To address these limitations, we propose using an exposure-based fairness evaluation metric previously applied in information retrieval~\cite{draws2021assessing}. Our experiments show that several existing approaches~\cite{rahman_fairwalk_2019,dong_edits_2022,ling_learning_2023,dai_say_2021,wang_improving_2022} fail at achieving a good tradeoff between accuracy and fairness under this evaluation metric. This motivates the design of a new post-processing algorithm that can be combined with any existing link prediction model to generate fair outcomes based on the new metric, outperforming existing alternatives. 

We summarize the contributions of this work as follows: (1) we expose the limitations of demographic parity as a fairness metric in the context of link prediction; (2) we propose using an exposure-based fairness metric that overcomes the limitations of demographic parity and show that existing approaches for fair link prediction are ineffective under the new metric; and (3) we introduce \textbf{MORAL}, a post-processing algorithm that can de-bias the outputs of any link prediction model and achieves good accuracy vs. fairness tradeoffs under the new evaluation metric.

\section{Preliminaries}

\label{sec::problem_statement}

Let a graph $G = (V, E, S)$, $v\in V$ is the set of nodes, $(u, v) \in E$ the set of existing edges in the graph, and $S \in \{0, 1\}^{n}$ is the vector of sensitive attributes, where $s_v$ indicates the sensitive attribute of node $v$. In this work, we consider binary sensitive attributes for simplicity and the availability of datasets, but our analysis is also valid for categorical and/or multiple sensitive attributes. The binary sensitive attribute $S$ produces three subgroups of node pairs, which for the remainder of the paper we denote as $E_{s \text{-} {s'}} = \{ (u, v) \in V\times V\mid s_u=1, s_v=0 \}$, $E_{s \text{-} s} = \{ (u, v) \in V\times V\mid s_u=1, s_v=1 \}$, and $E_{s' \text{-} s'} = \{ (u, v) \in V\times V \mid s_u=0, s_v=0 \}$, for simplicity. We consider a link prediction classifier a score function $f(\cdot,\cdot)$ that maps a given input pair $(u, v)$ to a score $\hat{Y} \in [0, 1]$. We denote the scores of all candidate pairs as $R \in [0, 1]^{|C|}$, where $C$ is the set of candidate pairs. 

\begin{definition}[Demographic Parity - $\Delta_{DP}$]
    Let $\hat{Y}$ be the prediction of a binary classifier. Demographic parity is defined as: $|P(\hat{Y} \mid (u, v) \in E_{\text{intra}}) - P(\hat{Y} \mid (u, v) \in E_{\text{inter}})|.$
\end{definition}


    In this scenario, we can define the objective of fair link prediction with an output $R$ as $\min_{R} \lambda\mathscr{L}_A(R)$, where $\mathscr{L}_A$ is an accuracy loss (usually Binary Cross Entropy), subject to a constraint (or regularization term) based on a bias loss $\mathscr{L}_B(R) \leq \beta$. Previous works \cite{li_dyadic_2021,li_fairlp_2022,current_fairegm_2022} follow the fair graph representation learning community practice of defining the task of fair link prediction in a \textit{dyadic} framework. In particular, these works propose to divide pairs into intra-pairs ($E_{\text{intra}} = \{(u, v) \in E \mid \ENSNS \bigcup \ESS\}$) and inter-pairs ($E_{\text{inter}} = \{(u, v) \in E \mid \ESNS\}$), and are trained and evaluated considering demographic parity ($\Delta_{DP}$) or one of its surrogates as the main fairness metric ($\mathscr{L}_B \approx \Delta_{DP}$).

\section{Limitations of Demographic Parity for Fair Link Prediction}
\label{sec::limitations_of_dp}


Despite recent advances in fair graph learning, existing bias mitigation techniques for link prediction remain limited in three fundamental ways. First, many approaches rely on node representations from GNNs with constrained expressive power, restricting their ability to model the nuanced structural and demographic patterns necessary for fairness. Second, common fairness metrics such as demographic parity ($\Delta_{DP}$) operate under a dyadic assumption that aggregates across sensitive subgroups, masking imbalances that occur within groups. Finally, these metrics are often permutation-invariant and insensitive to ranking order, making them unsuitable for applications where exposure and position matter. We systematically analyze these limitations to motivate a new fairness criterion for link prediction.

\subsection{Expressive Power Impact on Fairness}




Fair link prediction methods typically fall into two categories: node-level and link-level approaches. Node-level methods assume that fair node representations—often learned via Graph Neural Networks (GNNs)—will naturally result in fair link predictions~\cite{li_dyadic_2021}. In contrast, link-level methods attempt to directly enforce fairness on edge predictions, either by learning fairness-aware edge embeddings or by modifying the graph structure (e.g., through unbiased adjacency matrices). These methods often rely on adversarial training or fairness-specific loss functions to enforce demographic parity across intra- and inter-group links~\cite{li_dyadic_2021,current_fairegm_2022}. However, both approaches are fundamentally constrained by the limited expressive power of standard GNN architectures.

Most GNNs operate within the expressivity limits of the Weisfeiler-Lehman (1-WL) test, which restricts their ability to distinguish certain graph structures and node interactions \cite{xu2018powerful}. This poses a problem for fairness in link prediction, which often requires capturing subtle topological and demographic asymmetries across node pairs. For instance, a model must distinguish between link types (e.g., $\ESS$ vs. $\ESNS$) and detect systematic under-representation. Yet, 1-WL GNNs tend to produce similar embeddings for nodes in symmetric neighborhoods, failing to differentiate pairwise combinations that matter for fairness. See Appendix \ref{ap::limited_expressive_power} for an example of this limitation.


\subsection{Bias Within Edge Groups ($\ENSNS$ vs. $\ESS$)}
\label{subsec::aggegated_subgroups}
\begin{figure}[htbp]
    \centering
    \includegraphics[width=0.75\linewidth]{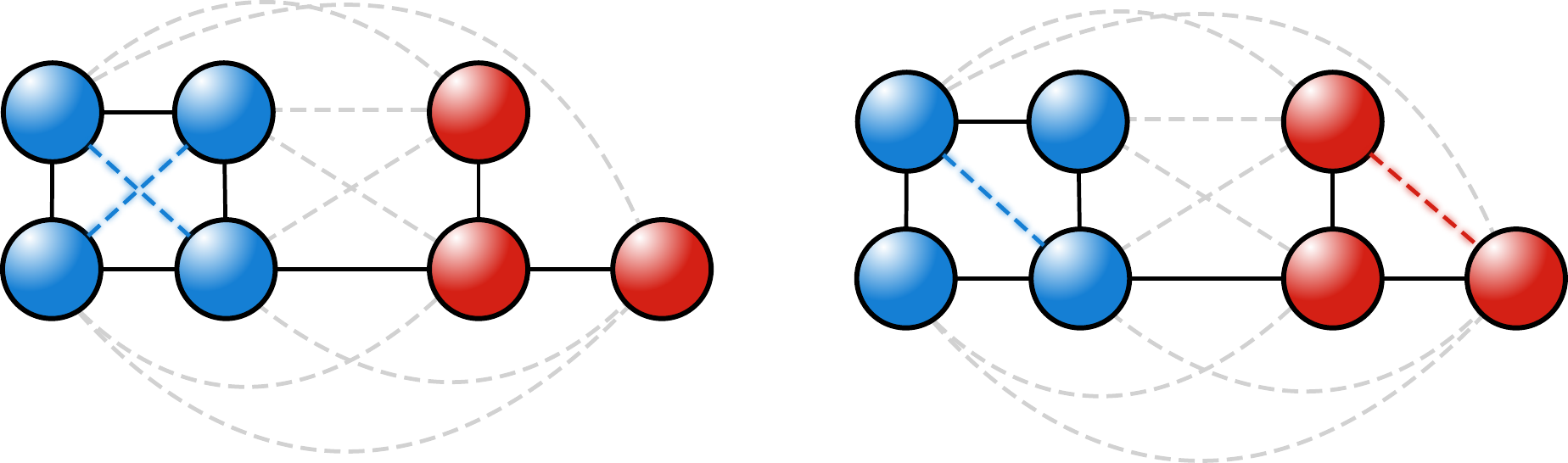}
    \caption{Toy example showing how \dempar fails to distinguish subgroup bias within aggregated edge groups. Both panels depict top-10 link predictions (dashed edges) over the same graph, achieving the optimal \dempar value despite the left scenario overrepresenting $\ENSNS$ (blue) relative to $\ESS$ (red), while $\ESNS$ (gray) is the protected group.}

    \label{fig:placeholder1}
\end{figure}

As discussed, the limited expressivity of node-level representations motivates the use of fair edge representations for link prediction. However, most existing methods aim to directly minimize \dempar, which can introduce unintended biases. Specifically, considering  \dempar at the edge level without accounting for the combinatorial nature of sensitive attributes can mask disparities between subgroup pairings.

In the case of a binary sensitive attribute (e.g., gender), the graph naturally decomposes into three edge types: $\ENSNS$, $\ESS$, and $\ESNS$, corresponding to all possible node pairings. Existing methods often aggregate these into broader categories—such as intra-group ($\ENSNS \cup \ESS$) and inter-group ($\ESNS$) edges—to enforce fairness constraints. However, we show that such aggregation may overlook systematic biases within the aggregated subgroups. For example, a model may consistently overpredict links of type $\ENSNS$ relative to $\ESS$, yet still satisfy \dempar under the aggregated grouping. More generally, the metric $\Delta_{\text{max}} = \max_{g_1, g_2 \in \mathcal{G}, g_1 \ne g_2} |P(\hat{Y} \mid g_1) - P(\hat{Y} \mid g_2)|$ fails to penalize subgroup-level disparities when groupings $\mathcal{G}$ ignore pairwise composition. As a result, traditional dyadic fairness assumptions break down in the link prediction setting.

\paragraph{\emph{Example}} Figure~\ref{fig:placeholder1} demonstrates how demographic parity can obscure subgroup imbalances. Here, $\ESNS$ (gray) represents inter-group links (designated as the protected class), while $\ENSNS$ (blue) and $\ESS$ (red) represent intra-group links. Although the model disproportionately favors $\ENSNS$ over $\ESS$ in the left panel, both panels yield identical \dempar scores due to aggregation into $E_{\text{intra}}$. Even alternative groupings (e.g., $\ENSNS \cup \ESNS$ vs. $\ESS$) fail to resolve the issue: underexposure of certain subgroups remains undetected. 

\paragraph{\emph{Toward Subgroup-Sensitive Fairness.}} Over multiple prediction rounds, this skew can amplify structural disparities, akin to exposure bias~\cite{singh_fairness_2018}, reinforcing phenomena such as the glass ceiling effect~\cite{stoica2018algorithmic}. To address this issue, we propose a fairness criterion that preserves the distributional structure of edge types as observed in the original graph.

\begin{property}[Non-Dyadic Distribution-Preserving Fairness]
    A fairness metric should treat all sensitive attribute pairings as distinct subgroups, and aim to align the predicted edge distribution with that of the original graph. Formally, let $\boldsymbol{\pi} = (\pi_{s\text{-}s}, \pi_{s'\text{-}s}, \pi_{s'\text{-}s'})$ denote the empirical distribution of edge types defined by a binary sensitive attribute $S$. Let $\hat{\boldsymbol{\pi}} = (\hat{\pi}_{s\text{-}s}, \hat{\pi}_{s'\text{-}s}, \hat{\pi}_{s'\text{-}s'})$ be the predicted distribution. Then, a fair link predictor should minimize $dist(\hat{\boldsymbol{\pi}}, \boldsymbol{\pi})$, where $dist$ is a suitable divergence metric (e.g., KL divergence).
    \label{property:non_dyadic}
\end{property}


\subsection{Bias Across Edge Groups ($E_{\text{intra}}$ vs. $E_{\text{inter}}$)}



\begin{figure}[t]
    \centering
    \includegraphics[width=1\linewidth]{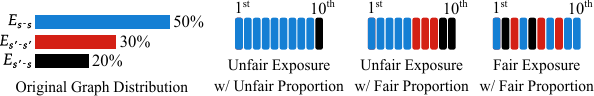}
        \caption{\dempar fails to capture exposure and subgroup proportion bias. Three models (a)–(c) output top-$10$ link predictions over a graph with original subgroup edge proportions of 50\% $\ENSNS$ (blue), 30\% $\ESS$ (red), and 20\% $\ESNS$ (black).}

    \label{fig:example2}
\end{figure}

Fair link prediction methods commonly frame the task as binary classification, using utility metrics such as AUC-ROC to evaluate model performance~\cite{li_fairlp_2022,li_dyadic_2021,current_fairegm_2022,masrour_bursting_2020}. However, this evaluation protocol often misaligns with real-world applications, where link prediction decisions are based on ranked candidate lists~\cite{tsioutsiouliklis_fairness-aware_2021, tsioutsiouliklis2022link}. Recent work addresses this limitation using ranking-based evaluation metrics \cite{mattos2025attribute}. 

In terms of fairness, many approaches adopt the \dempar reduction between \(E_{\text{inter}}\) and \(E_{\text{intra}}\) as the primary fairness metric, which does not consider the \textit{proportion} of pairs from each group in the final ranking. As a result, a biased model could produce a ranking heavily dominated by one pair type (e.g., \(E_{\text{intra}}\)) and still be evaluated as fair if the predicted scores are numerically similar. We argue that \dempar is inherently ranking-insensitive, making it an inadequate fairness measure for ranking-based tasks such as link prediction.

Even if the final ranking distribution of pairs of each type in a ranking approximates the distribution from the original graph, \dempar is also limited by being permutation invariant. This characteristic also enables another instance of exposure bias \cite{singh_fairness_2018}, but this time in a dyadic fashion. For instance, a (biased) link prediction algorithm can output a ranking of pairs that promotes an unfair number of $E_{\text{intra}}$ ($\ENSNS \bigcup \ESS$) pairs to the top positions while still maintaining low values of demographic parity, characterizing a case of exposure disparity against $E_{\text{inter}}$. This is problematic in many applications, in which user attention is concentrated on the few top-ranked items.



\paragraph{\emph{Example}} Figure~\ref{fig:example2} presents three hypothetical top-$10$ rankings produced by different link prediction models, each evaluated over the same input graph with subgroup edge distribution $\boldsymbol{\pi} = (0.5, 0.2, 0.3)$. Model (a) exhibits severe bias by over-representing $\ENSNS$ pairs, distorting both the group proportions and their relative exposure in top ranks. Model (b) preserves the global edge distribution but places $E_{\text{intra}}$ edges disproportionately high, resulting in exposure bias despite matching overall proportions. Model (c) maintains both proportionality and fair exposure across groups. Despite these clear disparities, all three rankings could yield the same \dempar score, which is blind to ranking order and group-specific position effects—highlighting the need for rank-aware fairness evaluation.







\begin{property}[Rank awareness]
    A fairness metric for link prediction should be sensitive to the proportion and rank of every type of pair. Specifically, it should ensure that any group, when ranked, does not systematically have higher or lower ranks compared to other groups. Let $\hat{\boldsymbol{\pi}}_k$ denote the distribution of each pair type in the top-$k$ ranking of pairs outputted by a link prediction classifier, and $\mathcal{C}$ the set of candidate pairs. A rank-aware fairness metric should minimize $\min_{\hat{\boldsymbol{\pi}}_k}\sum_{k=1}^{|\mathcal{C}|} dist(\hat{\boldsymbol{\pi}}_k,\boldsymbol{\pi}){\delta_k}$ where $\delta_k$ denotes the proportional exposure decay attributed to the $k$-th position on the ranking, which is usually monotonically decreasing. 
    \label{property:rank_awareness}
    
\end{property}

Previous definitions of group exposure \cite{singh_fairness_2018,zehlike_reducing_2020} consider the sum of the scores of items from the group weighted by the top-1 position bias. Such a definition assumes that items in top positions receive more attention \cite{joachims_accurately_nodate} and, thus, should have larger weights associated with their scores. Following the same assumption, Property \ref{property:rank_awareness} ensures that significant deviations from the original graph distribution in the top positions are more heavily penalized than deviations occurring at the bottom positions.

\section{Distribution-Preserving and Ranking-Aware Fair Link Prediction}

We first demonstrate how fair ranking evaluation metrics can address previous limitations associated with \dempar, and then we propose a simple post-processing algorithm for bias mitigation in link prediction methods.

\subsection{Fair Ranking Metrics}






The limitations of \dempar suggest the need for accounting not only for group proportions but also for exposure in ranked outputs. Inspired by previous works on fair ranking \cite{zehlike_fair_2017}, we pose fair link prediction as a group-aware ranking problem over candidate links.

In standard fair ranking, items are ranked by relevance, and fairness is enforced by controlling the representation of protected groups in top positions. Although link prediction differs from classical ranking—most notably, edges do not carry scalar relevance scores—we observe that a model still induces a ranking over predicted edges. Moreover, this induced ranking affects which types of node pairs (e.g., $E_{s\text{-}s}$, $E_{s'\text{-}s'}$) are prioritized in downstream model decisions.

Our setting breaks away from dyadic fairness assumptions by considering group-level edge categories as ranking units. Given this framing, we seek metrics and methods that jointly account for (i) consistency with group-level edge proportions and (ii) equitable exposure across ranking positions.

In this fashion, we adopt the \emph{Normalized Cumulative KL-Divergence} (NDKL) as our fairness metric. NDKL penalizes deviation between the cumulative exposure of group categories in the top-$k$ ranked edges and their expected distribution, capturing exposure fairness and subgroup proportions.

\begin{definition}[Normalized Cumulative KL-Divergence - NDKL]
    Let $\mathcal{C}$ be the set of candidate pairs ranked by score, $\boldsymbol{\pi}$ the original proportion of each sensitive edge group in the graph, and $\hat{\boldsymbol{\pi}}_k$ the distribution of sensitive groups up to the $k$-th position of $\mathcal{C}$, the NDKL is:
    
    \[
    \mathrm{NDKL} = \frac{1}{Z} \sum_{k=1}^{|\mathcal{C}|} \frac{1}{\log_2(k+1)}\, D_{\mathrm{KL}}\left( \hat{\boldsymbol{\pi}}_k \,\|\, \boldsymbol{\pi} \right),
    \]
where $D_{\mathrm{KL}}(p||q)$ is the KL-Divergence between distributions $p$ and $q$, and $Z=\sum_{i=1}^{|\mathcal{C}|}\frac{1}{\log_2(i+1)}$ is a normalizer.
\end{definition}

\begin{theorem}
    Let $\boldsymbol{\pi} = [\pi_0, \pi_1, \pi_2]$ be the target distribution over $3$ sensitive groups, with $\sum_i \pi_i = 1$, and let $\hat{\boldsymbol{\pi}}_k$ denote the empirical distribution over the top-$k$ ranked items. Under the constraint that the full ranking satisfies demographic parity (i.e., the overall empirical distribution matches $\boldsymbol{\pi}$), the NDKL score satisfies the following bounds: $0 \;\leq\; \mathrm{NDKL} \;\leq\; \max_{i \in \{0,1,2\}} \log \frac{1}{\pi_i}$.
\label{theorem::ndkl}
\end{theorem}

Theorem~\ref{theorem::ndkl} establishes upper and lower bounds for the NDKL score under the assumption of demographic parity at the full-ranking level. To empirically validate this result, we construct controlled scenarios where candidate rankings are manipulated to explore different levels of exposure bias—ranging from completely fair to maximally skewed under a fixed $\boldsymbol{\pi}$. As shown in Section~\ref{subsec::dem_par_gap}, the observed NDKL scores in these settings consistently lie within the theoretical bounds, confirming the expected behavior by responding sensitively to violations in exposure fairness.

\color{black}


NDKL satisfies both key properties required for fairness in link prediction. By using KL-Divergence over group distributions, it supports multiple sensitive groups without requiring dyadic aggregation (Property~\ref{property:non_dyadic}). It is also rank-aware, which ensures top-ranked exposures are more influential (Property~\ref{property:rank_awareness}). While NDKL was previously used in fair ranking~\cite{geyik_fairness-aware_2019} and is our metric of choice, other metrics could be used—provided they quantify divergence across multiple groups and incorporate exposure weighting in ranked outputs.

\subsection{MORAL - Multi-Output Ranking Aggregation for Link Fairness}

We introduce \textbf{MORAL} (Multi-Output Ranking Aggregation for Link fairness), a simple and scalable post-processing framework designed to improve fairness in link prediction. MORAL decouples group-wise predictions and enforces exposure parity through a ranking aggregation mechanism.

Specifically, MORAL trains three distinct link prediction models: $f_{s\text{-}s}$, $f_{s\text{-}s'}$, and $f_{s'\text{-}s'}$, each trained exclusively on edges corresponding to a specific sensitive group interaction. This decoupling mitigates group-specific utility disparities and prevents a single model from exhibiting imbalanced predictive performance across groups. In addition, MORAL remains computationally efficient even for sensitive attributes with larger cardinality, as each model processes only $\frac{|E_{train}|}{|S|\cdot{|S|\choose2}\cdot b}$ gradients per epoch, where $|E_{train}|$ denotes the number of training edges, $|S|$ the number of sensitive attribute categories, and $b$ the batch size.
\color{black}

At inference time, MORAL aggregates predictions from the group-specific models into a unified ranking. This is accomplished by maintaining a running estimate of the exposure distribution across the three edge types and greedily selecting, at each rank position, the highest-scoring remaining edge from the model whose inclusion most reduces the cumulative KL divergence from a predefined target distribution $\boldsymbol{\pi}$ (see Algorithm~\ref{alg:greedy_dkl} for pseudocode). This exposure-aware ranking procedure ensures that the final output approximates the desired group-wise exposure proportions.

Our greedy strategy solves the following fairness-prioritized objective:
$\min_{R} \mathscr{L}_A(R) \quad \text{s.t.} \quad \mathscr{L}_B(R) \leq \min_{R'} \mathscr{L}_B(R')$. Similarly, one could optimize a different objective that balances accuracy and fairness through a hyperparameter $\lambda$. A greedy algorithm can optimize this weighted-sum objective by minimizing the combined loss at each step \cite{celis2018ranking}. We opt for the constrained formulation above to explicitly prioritize fairness across all datasets, especially in imbalanced settings where exposure risks are more severe.

MORAL addresses the challenges identified in Section~\ref{sec::limitations_of_dp} by combining group-specific models with KL-guided ranking. Moreover, MORAL outputs group assignments per rank position, meaning it can be seamlessly paired with any fair ranking metric (like NDKL) that satisfies Properties~\ref{property:non_dyadic} and~\ref{property:rank_awareness}.






\begin{algorithm}[H]
\caption{MORAL: Multi-Output Ranking Aggregation for Link Fairness}
\label{alg:greedy_dkl}
\KwIn{
    \begin{itemize}
        \item Candidate sets $\mathcal{C}_j = \{(u,v, \text{score})\}$ for each group $j \in \{0,1,2\}$ (sorted by descending score);
        \item Target distribution $\boldsymbol{\pi} = (\pi_0, \pi_1, \pi_2)$;
        \item Total output size $n$.
    \end{itemize}
}
\KwOut{Ranking list $\mathbf{R}$ of $n$ predicted edges with assigned group labels}

Initialize exposure counts: $\boldsymbol{c} \leftarrow (0, 0, 0)$\;
Initialize output ranking: $\mathbf{R} \leftarrow [\;]$\;

\For{$t \leftarrow 1$ \KwTo $n$}{
    Initialize best objective: $\text{min\_kl} \leftarrow \infty$, $\text{selected\_group} \leftarrow -1$, $\text{selected\_edge} \leftarrow \text{None}$\;

    \ForEach{group $j \in \{0,1,2\}$ such that $\mathcal{C}_j$ is not empty}{
        Let $(u,v, \text{score}) \leftarrow$ top element in $\mathcal{C}_j$\;
        
        Temporarily update counts: $c_j' \leftarrow c_j + 1$, $q_j' \leftarrow \frac{c_j'}{t}$, $q_{j'\neq j} \leftarrow \frac{c_{j'}}{t}$\;
        
        Compute KL divergence: $D_{\mathrm{KL}}( \mathbf{q}' \| \boldsymbol{\pi} )$\;
        
        \If{this KL is lower than $\text{min\_kl}$}{
            Update $\text{min\_kl} \leftarrow D_{\mathrm{KL}}$, $\text{selected\_group} \leftarrow j$, $\text{selected\_edge} \leftarrow (u,v)$\;
        }
    }

    Append $(\text{selected\_edge}, \text{selected\_group})$ to $\mathbf{R}$\;
    
    Remove top element from $\mathcal{C}_{\text{selected\_group}}$\;
    
    Update $c_{\text{selected\_group}} \leftarrow c_{\text{selected\_group}} + 1$\;
}

\Return{$\mathbf{R}$}
\end{algorithm}

\section{Experiments}
\label{sec::experiments}

\begin{table}[htbp]
\centering
\resizebox{\columnwidth}{!}{%
\begin{tabular}{@{}lcccccccc@{}}
\toprule
\textbf{Dataset} & \textbf{$|V|$} & \textbf{$|E|$} & \textbf{Feat.} & \textbf{Attr.} & \textbf{$\ESNS$ (\%)} & \textbf{$\ENSNS$ (\%)} & \textbf{$\ESS$ (\%)} & \textbf{Topo.} \\ \hline
facebook         & 1045           & 18726          & 573            & Gen.                          & 42            & 44             & 14           & Periph.        \\
german           & 1000           & 15220          & 27             & Age                           & 20            & 61             & 19           & Periph.        \\
nba              & 403            & 7435           & 95             & Nat.                          & 27            & 63             & 10           & Periph.        \\
pokec\_n         & 66569          & 361934         & 276            & Gen.                          & 5             & 66             & 29           & Comm.          \\
pokec\_z         & 67796          & 432572         & 265            & Gen.                          & 5             & 58             & 37           & Comm.          \\
credit           & 30000          & 96165          & 13             & Age                            & 12            & 86             & 2            & Periph.        \\ \bottomrule
\end{tabular}%
}
\caption{Statistics from our six real-world datasets, covering periphery and community graphs from different domains.}
\label{tab:dataset_stats}
\end{table}

We evaluate link prediction approaches across multiple datasets through fairness and link prediction metrics based on ranking. Considering our previous claims, we formulate three main research questions: \textit{ RQ1: To what extent does \dempar lead to hidden biases in the proportion of each group type in link prediction tasks?} \textit{RQ2: Does ranking-awareness lead to a more faithful assessment of fairness in link prediction, compared to dyadic or proportion-based measures?} \textit{RQ3: How does MORAL compare against existing fair link prediction approaches under the ranking metric?} 


\textbf{Baselines} To enable a robust comparison for fair link prediction, we evaluate a diverse and competitive set of methods: a node embedding approach (FairWalk ~\cite{rahman_fairwalk_2019}); pre-processing methods (EDITS ~\cite{dong_edits_2022}, FairLP ~\cite{li2022fairlp}, FairEGM ~\cite{current2022fairegm}); in-processing methods (GraphAIR ~\cite{ling_learning_2023}, UGE ~\cite{wang2022unbiased}); post-processing methods (DetConstSort ~\cite{geyik_fairness-aware_2019}, MORAL); and a task-specific fair link prediction method (FairAdj ~\cite{li_dyadic_2021}). For consistency, all approaches use a GCN encoder ($f_\theta$) with a dot-product decoder, following the framework in Section~\ref{sec::problem_statement}. We expect similar trends with other GNNs/SGNNs \cite{wang_neural_2024, zhang_link_2018, chamberlain2022graph}. Hyperparameters follow each method’s original recommendations.

\textbf{Datasets.} We conduct experiments on six real-world datasets considered in previous works \cite{dong_edits_2022,li_fairlp_2022,current_fairegm_2022}. These datasets represent diverse domains and fairness contexts. The Credit dataset is a credit scoring network where nodes represent individuals and edges represent credit relationships, with age as sensitive attribute. The Facebook dataset is a social network where nodes are users and edges represent friendships, with gender as sensitive attribute. The German dataset is a credit approval network where nodes are individuals and edges link similar individuals, with gender as sensitive attribute. The NBA dataset is a network of NBA basketball players, where edges represent relationships between the athletes on Twitter, and nationality ('US' and 'overseas') is the sensitive attribute. The Pokec-n/z datasets consist of all the data from Pokec, a social network from Slovakia, in 2012, where nodes are users, edges are friendships, and gender is the sensitive attribute. Table \ref{tab:dataset_stats} presents statistics for each dataset, including the number of nodes, edges, and the distribution of sensitive attributes. We selected datasets from two different graph distributions (community and periphery).


\textbf{Evaluation Metrics.} We consider two metrics, adhering to the evaluation scheme of ranking tasks. We adopt NDKL as our fairness ranking metric and \emph{prec@k} as our ranking-based performance metric. We also establish comparisons between NDKL and \dempar in Section \ref{subsec::dem_par_gap}.


\textbf{Implementation Details} We implement all models using PyTorch Geometric \cite{Fey/Lenssen/2019} and the PyGDebias \cite{dong2023fairness}. We run experiments using NVIDIA A40 GPUs, fixing random seeds, and run each experiment 3 times. Random edge splits are set as 70/10/20\% for training/validation/testing. The sensitive attribute distribution is preserved across all splits. We adopt the Adam optimizer with learning rate $0.0003$ for all experiments.


\subsection{Hidden Bias in Dyadic Fairness}

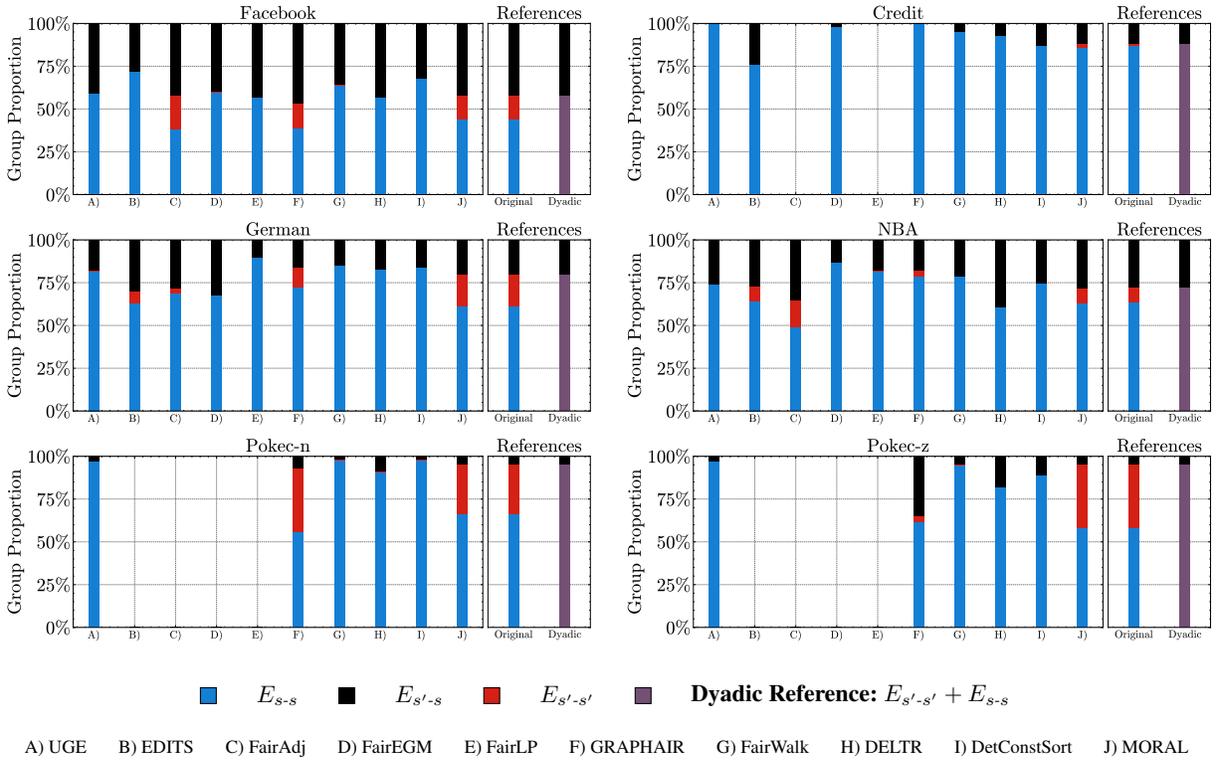
\begin{figure*}[h]
    \centering
    \begin{tabular}{cc}
        \includegraphics[width=0.44\textwidth]{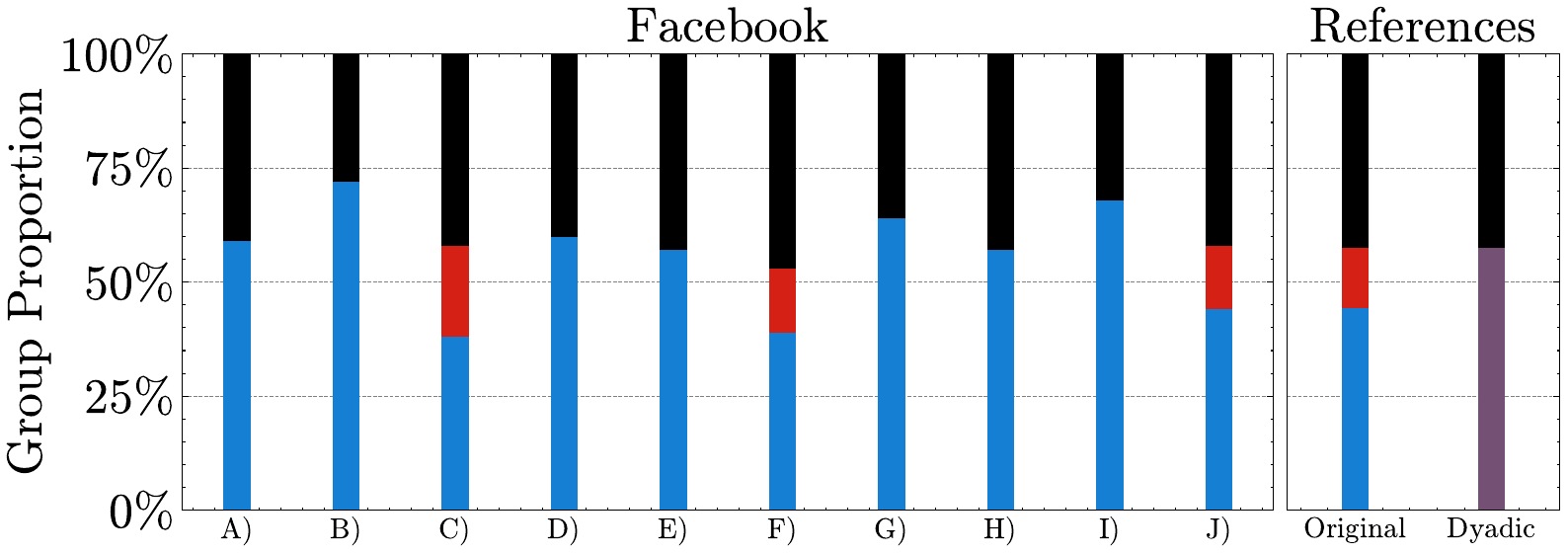} &
        \includegraphics[width=0.44\textwidth]{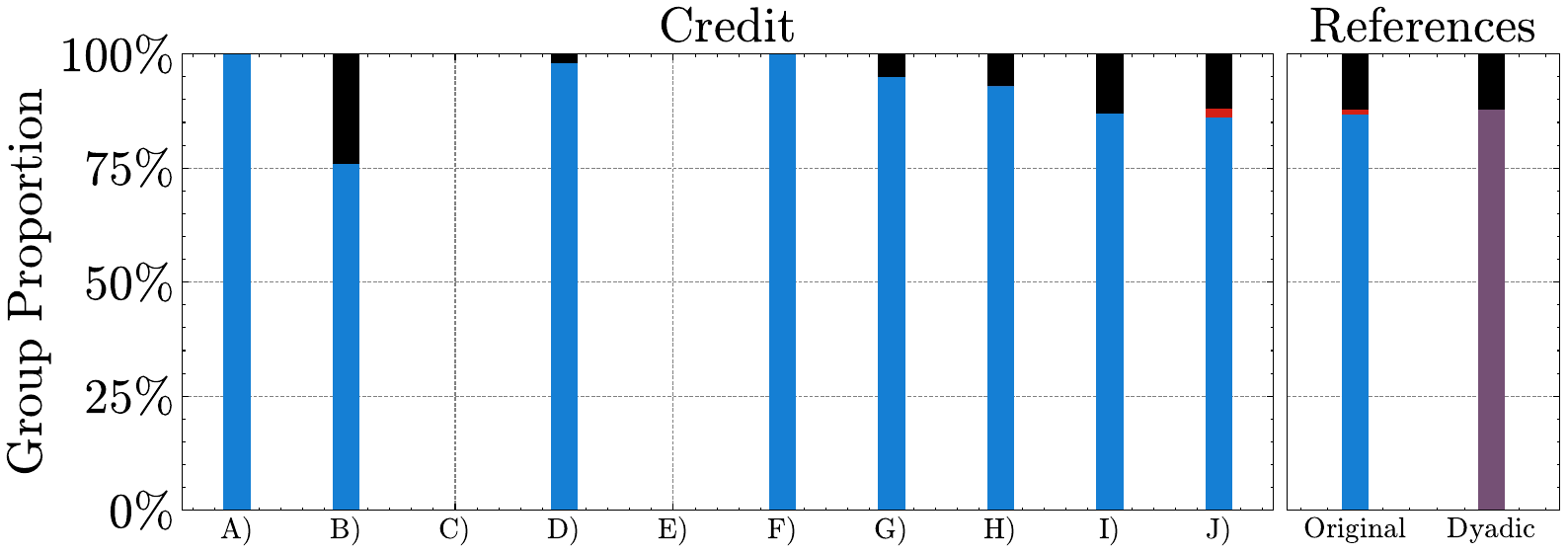} \\
        \includegraphics[width=0.44\textwidth]{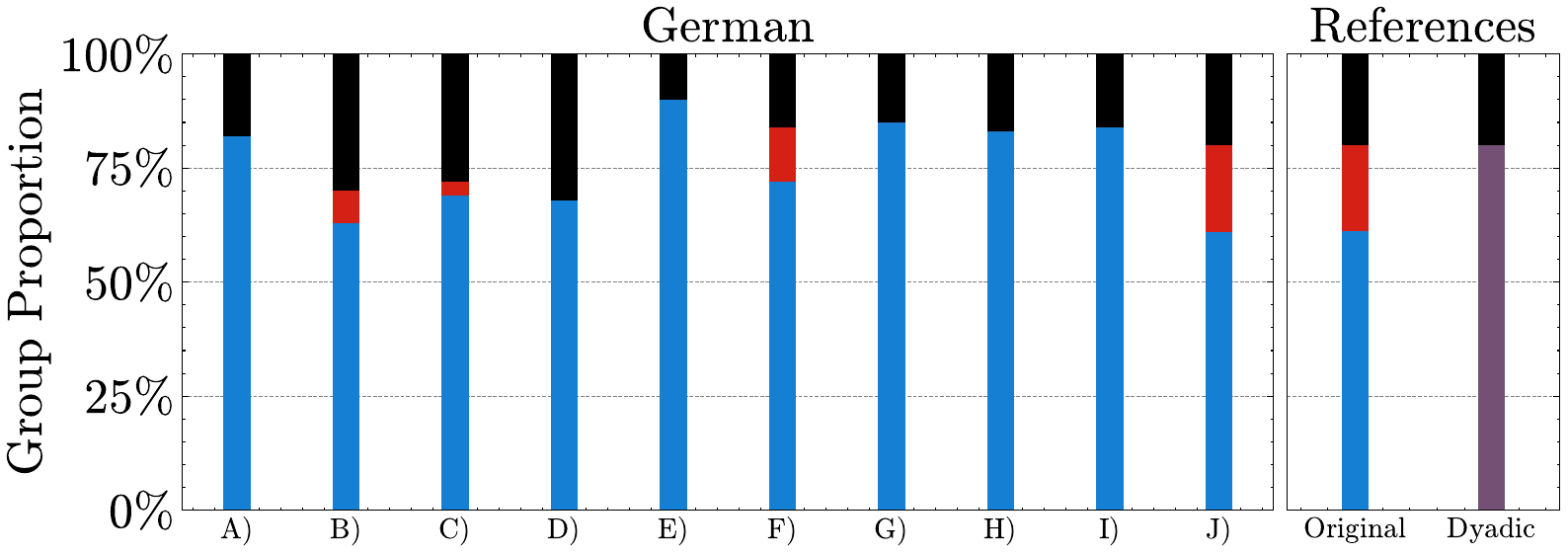} &
        \includegraphics[width=0.44\textwidth]{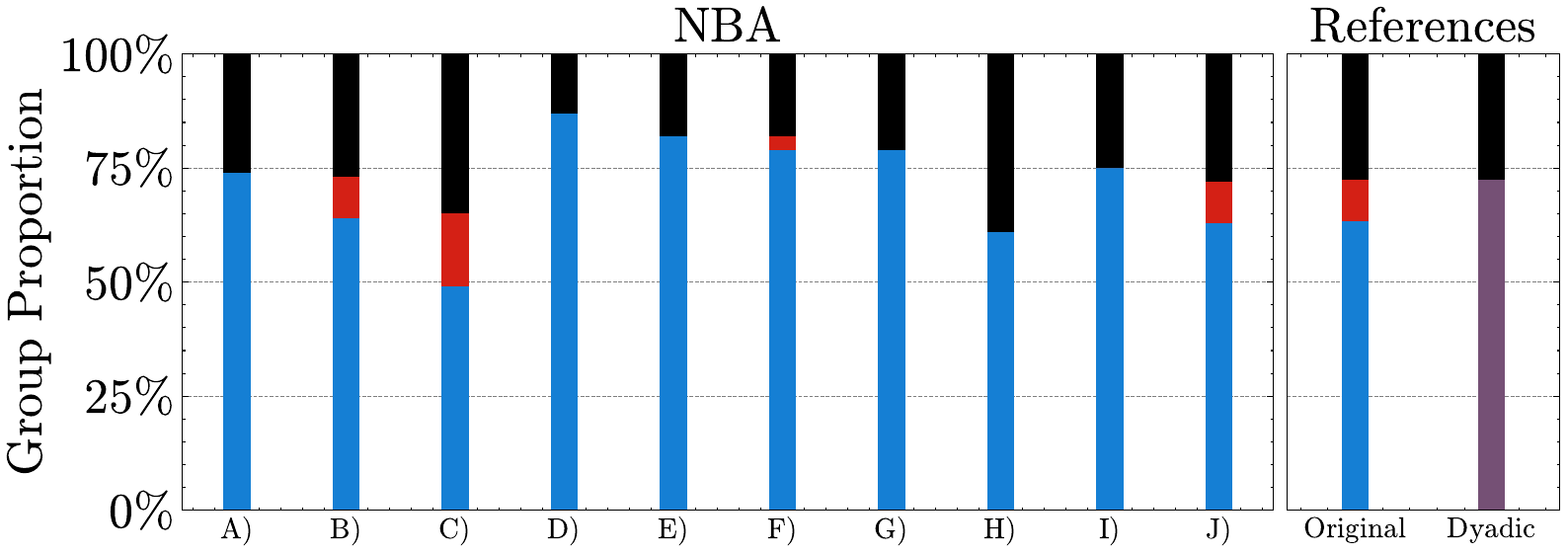} \\
        \includegraphics[width=0.44\textwidth]{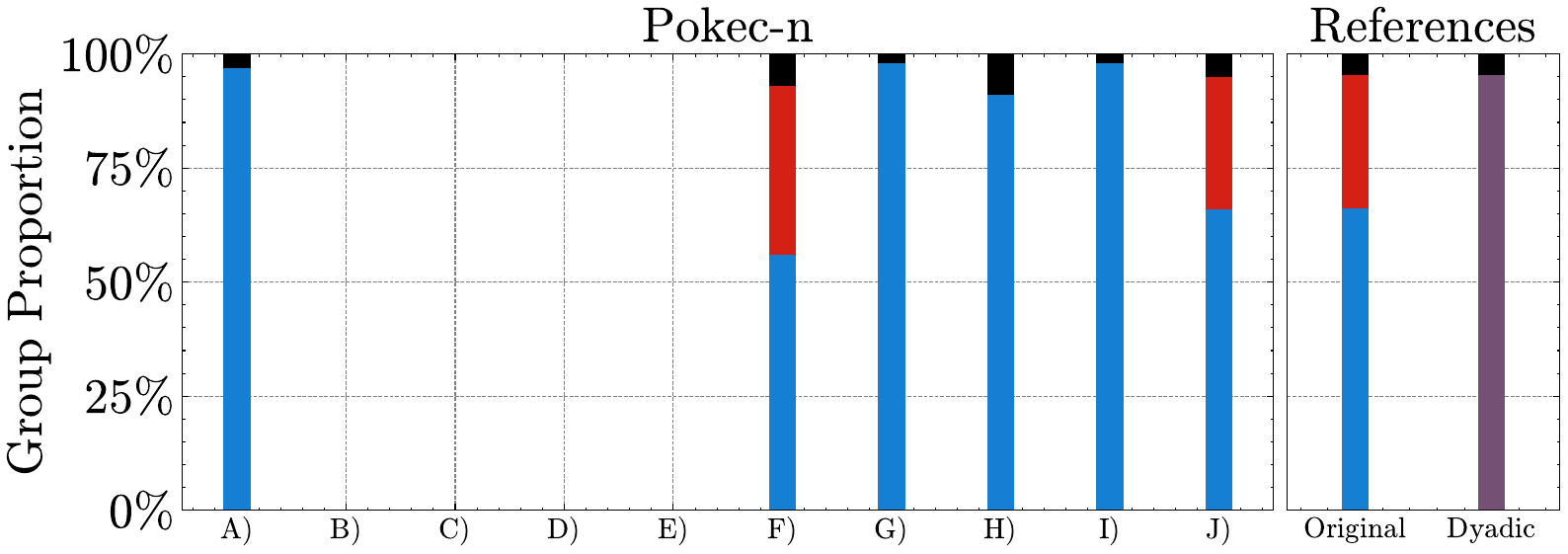} &
        \includegraphics[width=0.44\textwidth]{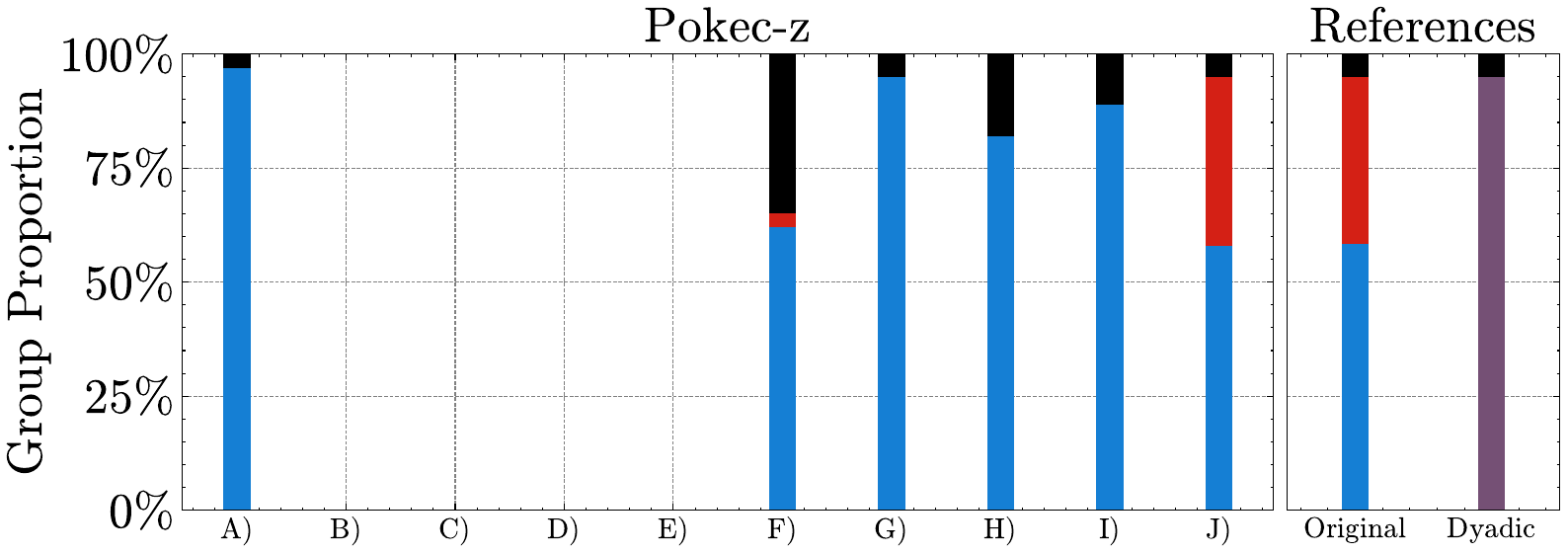} \\
    \end{tabular}

     \vspace{0.5em}

    \begin{tikzpicture}[scale=1, every node/.style={scale=0.9}]

        \matrix [row sep=0pt, column sep=12pt] {
            \node[draw=black, fill=blueish, minimum size=6pt] {}; & \node { $\ENSNS$ }; &
            \node[draw=black, fill=black, minimum size=6pt] {};  & \node { $\ESNS$ }; &
            \node[draw=black, fill=redish, minimum size=6pt] {}; & \node { $\ESS$ }; &
            \node[draw=black, fill=purpleish, minimum size=6pt] {}; & \node { \textbf{Dyadic Reference:} $\ESS + \ENSNS$ }; \\
        };
    \end{tikzpicture}

        {\scriptsize
    \begin{tabular}{cccccccccc}
        A) UGE & B) EDITS & C) FairAdj & D) FairEGM & E) FairLP & F) GRAPHAIR & G) FairWalk & H) DELTR & I) DetConstSort & J) MORAL \\
    \end{tabular}
    }


    \caption{Proportions of pair types in the top-$100$ predictions by method, compared against the original graph distribution and an optimal dyadic fairness reference. Colors: $\ENSNS$ (\textcolor{blueish}{\textbf{blue}}), $\ESNS$ (\textcolor{black}{\textbf{black}}), and $\ESS$ (\textcolor{redish}{\textbf{red}}). In the dyadic fairness reference, \textcolor{purpleish}{\textbf{purple}} represents the combined proportion of $\ESS$ and $\ENSNS$ pairs. Missing bars indicate an OOM error.}
    \label{fig:grid_plots}
\end{figure*}

\begingroup
\fontsize{9}{10}\selectfont     
\begin{table*}[!h]
\centering
\small
\setlength{\tabcolsep}{4pt}
\caption{Fairness performance comparison of all approaches considered ($k=1000$). Lower \textit{NDKL} and higher \textit{prec@1000} are better. Best NDKL and \textit{prec@1000} values are in \textbf{bold} and \underline{underline}, respectively.}
\begin{tabular}{llcccccc}
\toprule
\textbf{Method} &  & \textbf{Facebook} & \textbf{Credit} & \textbf{German} & \textbf{NBA} & \textbf{Pokec-n} & \textbf{Pokec-z} \\
\midrule
\multirow{2}{*}{UGE} & NDKL & 0.05 $\pm$ 0.00 & 0.80 $\pm$ 0.07 & 0.08 $\pm$ 0.03 & 0.07 $\pm$ 0.02 & 0.06 $\pm$ 0.00 & 0.06 $\pm$ 0.00 \\
 & $prec@1000$ & \underline{0.97 $\pm$ 0.00} & \underline{1.00 $\pm$ 0.00} & 0.69 $\pm$ 0.01 & 0.58 $\pm$ 0.01 & 0.90 $\pm$ 0.04 & 0.91 $\pm$ 0.04 \\

\midrule

\multirow{2}{*}{EDITS} & NDKL & 0.21 $\pm$ 0.07 & 0.04 $\pm$ 0.02 & 0.08 $\pm$ 0.02 & 0.08 $\pm$ 0.01 & OOM & OOM \\
 & $prec@1000$ & 0.96 $\pm$ 0.00 & 0.36 $\pm$ 0.17 & 0.42 $\pm$ 0.20 & 0.49 $\pm$ 0.00 & OOM & OOM \\
 
\midrule
\multirow{2}{*}{GRAPHAIR} & NDKL & 0.13 $\pm$ 0.03 & 0.67 $\pm$ 0.22 & 0.07 $\pm$ 0.02 & 0.09 $\pm$ 0.01 & 0.09 $\pm$ 0.03 & 0.26 $\pm$ 0.25 \\
 & $prec@1000$ & 0.96 $\pm$ 0.01 & \underline{1.00 $\pm$ 0.00} & 0.73 $\pm$ 0.01 & 0.69 $\pm$ 0.01 & 0.97 $\pm$ 0.03 & \underline{1.00 $\pm$ 0.00} \\

\midrule

 \multirow{2}{*}{FairEGM} & NDKL & 0.09 $\pm$ 0.01 & 0.11 $\pm$ 0.00 & 0.05 $\pm$ 0.01 & 0.07 $\pm$ 0.01 & OOM & OOM \\
        & $prec@1000$ & 0.97 $\pm$ 0.00 & 1.00 $\pm$ 0.00 & 0.62 $\pm$ 0.00  & 0.60 $\pm$ 0.01 & OOM & OOM \\

\midrule

\multirow{2}{*}{FairLP} & NDKL & 0.18 $\pm$ 0.00 & OOM & 0.06 $\pm$ 0.00   & 0.20 $\pm$ 0.00 & OOM & OOM \\
        & $prec@1000$ &  0.99 $\pm$ 0.00 & OOM & 0.97 $\pm$ 0.00 & 0.86 $\pm$ 0.00 & OOM & OOM \\
 
\midrule
\multirow{2}{*}{FairWalk} & NDKL & 0.06 $\pm$ 0.01 & 0.06 $\pm$ 0.03 & 0.11 $\pm$ 0.02 & 0.06 $\pm$ 0.01 & 0.07 $\pm$ 0.01 & 0.07 $\pm$ 0.00 \\
 & $prec@1000$ & 0.96 $\pm$ 0.00 & \underline{1.00 $\pm$ 0.00} & 0.94 $\pm$ 0.00 & 0.55 $\pm$ 0.01 & \underline{1.00 $\pm$ 0.00} & \underline{1.00 $\pm$ 0.00} \\
 
\midrule
\multirow{2}{*}{FairAdj} & NDKL & 0.10 $\pm$ 0.05 & OOM & 0.10 $\pm$ 0.01 & 0.11 $\pm$ 0.05 & OOM & OOM \\
 & $prec@1000$ & 0.42 $\pm$ 0.01 & OOM & 0.54 $\pm$ 0.01 & 0.50 $\pm$ 0.01 & OOM & OOM \\
 
\midrule
\multirow{2}{*}{DetConstSort} & NDKL & 0.15 $\pm$ 0.00 & 0.06 $\pm$ 0.00 & 0.04 $\pm$ 0.00 & 0.09 $\pm$ 0.00 & 0.07 $\pm$ 0.00 & 0.23 $\pm$ 0.00 \\
 & $prec@1000$ & 0.00 $\pm$ 0.00 & 0.00 $\pm$ 0.00 & 0.55 $\pm$ 0.00 & 0.21 $\pm$ 0.00 & 0.07 $\pm$ 0.00 & 0.01 $\pm$ 0.00 \\
\midrule
\multirow{2}{*}{DELTR} & NDKL & 0.10 $\pm$ 0.03 & 0.03 $\pm$ 0.00 & 0.09 $\pm$ 0.06 & 0.09 $\pm$ 0.02 & 0.23 $\pm$ 0.23 & 0.22 $\pm$ 0.20 \\
 & $prec@1000$ & 0.91 $\pm$ 0.05 & 0.56 $\pm$ 0.29 & 0.31 $\pm$ 0.44 & 0.43 $\pm$ 0.24 & 0.65 $\pm$ 0.01 & 0.48 $\pm$ 0.28 \\
\midrule
\multirow{2}{*}{MORAL} & NDKL & \textbf{0.04 $\pm$ 0.00} & \textbf{0.01 $\pm$ 0.00} & \textbf{0.03 $\pm$ 0.00} & \textbf{0.02 $\pm$ 0.00} & \textbf{0.03 $\pm$ 0.00} & \textbf{0.04 $\pm$ 0.00} \\
 & $prec@1000$ & 0.95 $\pm$ 0.01 & \underline{1.00 $\pm$ 0.00} & \underline{0.96 $\pm$ 0.00} & \underline{0.80 $\pm$ 0.00} & 0.98 $\pm$ 0.00 & \underline{0.98 $\pm$ 0.00} \\
\bottomrule
\end{tabular}
\label{tab:results}
\end{table*}
\endgroup

\label{subsec::hidden_bias}

We first aim to answer RQ1, demonstrating the limitations of dyadic fairness for link prediction. We analyze the distributions of types of pairs in the top-$k$ of each method, independent of the order of the elements. In Figure \ref{fig:grid_plots}, we compare the proportions of pairs in the top-$k$ ($k=100$) of each baseline. Our approach obtains the closest approximation to the target distribution, while other methods overestimate one pair type at the detriment of another, despite obtaining low values of \dempar. This result demonstrates the limitation of \dempar in detecting underrepresented aggregated groups, and sheds light on the importance of adhering to Property \ref{property:non_dyadic}. 


\subsection{Demographic Parity Gap}
\label{subsec::dem_par_gap}

To answer RQ2, we demonstrate the effect of ranking on fairness in link prediction by fixing the proportion of each pair type. First, we compute the required proportions of each pair type for optimal \dempar. Then, we fix these proportions, but consider the worst and best possible permutations of these candidate pairs in terms of NDKL. 
To isolate the fairness component from the utility measurement, we assume in this experiment that the pairs being ranked are all positive, meaning that regardless of the permutation of the pairs, both rankings will obtain $prec@k=100\%$. We expose the NDKL gap for varying-sized rankings in the Appendix \ref{ap::ndkl_experimental_bounds}. Despite relevant exposure bias between the worst and best rankings, the value of \dempar is the same. This indicates the necessity of incorporating Property \ref{property:rank_awareness} into fair link prediction metrics.

\subsection{Baselines Ranking Comparison}
\label{subsec::baselines_ranking_comparison}

We compare our method against UGE, EDITS, FairAdj, GRAPHAIR, FairWalk, DELTR, and DetConstSort. 
We adopt \textit{NDKL} as the fairness metric and \textit{prec@1000} as the utility metric. For EDITS, we use the dot product between the node embeddings obtained by training a GCN on the fair graph generated as the pair scores. For GRAPHAIR, UGE, and FairWalk, the scores are obtained through the dot product of the fair node embeddings outputted. 

We show our results in Table \ref{tab:results}. MORAL achieves the fairest rankings while still maintaining high link prediction performance. Both periphery and community graph types are challenging for all methods. We highlight the large fairness improvements obtained by our approach across all datasets, particularly in Credit, German, and NBA.

\section{Related Work}
\label{sec::related_work}

We situate MORAL within two major areas: fair graph representation learning and fair ranking. Our approach is an example of a post-processing method, which has proven effective in fair ranking tasks \cite{xian2023fair,tifrea2023frappe,gorantla_problem_2021,zehlike_fair_2017,li2021user}, and aligns with decoupling classifiers for group fairness, which has been successfully explored in the past \cite{dwork_decoupled_2018}.

\subsection{Fair Graph Representation Learning}

\textbf{Pre-processing.} Methods such as EDITS \cite{dong_edits_2022} and ALFR \cite{edwards2015censoring} mitigate bias before training by modifying node features or the graph structure. Despite their effectiveness, these methods can be computationally expensive and face scalability issues.

\textbf{In-processing.} FairGNN \cite{dai_say_2021}, NIFTY \cite{agarwal_towards_2021}, GRAPHAIR \cite{ling_learning_2023}, FairVGNN \cite{wang_improving_2022} , and UGE \cite{wang2022unbiased} integrate fairness during training using adversarial learning or graph augmentations. These methods assume fairness can be enforced at the node embedding level, which may encounter expressivity power limitations, and not generalize to pairwise tasks like link prediction.

\textbf{Fair Embeddings.} FairWalk \cite{rahman_fairwalk_2019} and RELIANT \cite{dong_reliant_2023} focus on debiasing unsupervised embeddings via neighbor sampling and proxy removal, respectively. However, these methods rely heavily on the structural properties of the graph and are not designed for post-hoc ranking fairness.

\subsection{Fair Ranking and Information Retrieval}

Post-processing approaches like FA*IR \cite{zehlike_fair_2017} and DetConstSort \cite{geyik_fairness-aware_2019} re-rank outputs to satisfy fairness constraints. Though model-agnostic, they can be computationally intensive and inflexible. In-processing ranking methods, such as DELTR \cite{zehlike_reducing_2020}, policy learning \cite{singh_policy_2019,yadav_policy-gradient_2021}, and constrained optimization via SPOFR \cite{kotary_end--end_2021}, jointly optimize fairness and relevance. However, they typically target dyadic ranking tasks in learn-to-rank settings, which is misaligned with the structural nature of the link prediction task considered in this work.

\section{Conclusion}

Fairness in link prediction is a relevant problem addressed by a diverse set of previous approaches in the literature. In this work, we scrutinize the main assumptions behind how previous works evaluate the fairness of link prediction models. In particular, we shed light on pitfalls related to naively adopting dyadic fairness notions for link prediction and how this approach is prone to a hidden form of bias within aggregated subgroups. Further, we demonstrate how not capturing ranking notions in the fairness metric can potentially prevent a given metric from capturing exposure bias.


\section*{Acknowledgements}
We acknowledge the support by the US Department of Transportation Tier-1 University Transportation Center (UTC) Transportation Cybersecurity Center for Advanced Research and Education (CYBER-CARE) (Grant No. 69A3552348332), and the Rice Ken Kennedy Institute.

\bibliography{references}

@inproceedings{li2022fairlp,
  title={Fairlp: Towards fair link prediction on social network graphs},
  author={Li, Yanying and Wang, Xiuling and Ning, Yue and Wang, Hui},
  booktitle={Proceedings of the international AAAI conference on web and social media},
  volume={16},
  pages={628--639},
  year={2022}
}

@inproceedings{current2022fairegm,
  title={Fairegm: fair link prediction and recommendation via emulated graph modification},
  author={Current, Sean and He, Yuntian and Gurukar, Saket and Parthasarathy, Srinivasan},
  booktitle={Proceedings of the 2nd ACM Conference on Equity and Access in Algorithms, Mechanisms, and Optimization},
  pages={1--14},
  year={2022}
}

@inproceedings{wang2022unbiased,
  title={Unbiased graph embedding with biased graph observations},
  author={Wang, Nan and Lin, Lu and Li, Jundong and Wang, Hongning},
  booktitle={WebConf},
  year={2022}
}

@article{dai2024comprehensive,
  title={A comprehensive survey on trustworthy graph neural networks: Privacy, robustness, fairness, and explainability},
  author={Dai, Enyan and Zhao, Tianxiang and Zhu, Huaisheng and Xu, Junjie and Guo, Zhimeng and Liu, Hui and Tang, Jiliang and Wang, Suhang},
  journal={Machine Intelligence Research},
  volume={21},
  number={6},
  pages={1011--1061},
  year={2024},
  publisher={Springer}
}

@inproceedings{li2021user,
  title={User-oriented fairness in recommendation},
  author={Li, Yunqi and Chen, Hanxiong and Fu, Zuohui and Ge, Yingqiang and Zhang, Yongfeng},
  booktitle={WebConf},
  year={2021}
}

@inproceedings{luo2023cross,
  title={Cross-links matter for link prediction: rethinking the debiased GNN from a data perspective},
  author={Luo, Zihan and Huang, Hong and Lian, Jianxun and Song, Xiran and Xie, Xing and Jin, Hai},
  booktitle={NeurIPS},
  year={2023}
}

@inproceedings{laclau2021all,
  title={All of the fairness for edge prediction with optimal transport},
  author={Laclau, Charlotte and Redko, Ievgen and Choudhary, Manvi and Largeron, Christine},
  booktitle={AISTATS},
  year={2021},
}

@inproceedings{stoica2024fairness,
  title={Fairness rising from the ranks: HITS and pagerank on homophilic networks},
  author={Stoica, Ana-Andreea and Litvak, Nelly and Chaintreau, Augustin},
  booktitle={WebConf},
  year={2024}
}

@inproceedings{stoica2018algorithmic,
  title={Algorithmic glass ceiling in social networks: The effects of social recommendations on network diversity},
  author={Stoica, Ana-Andreea and Riederer, Christopher and Chaintreau, Augustin},
  booktitle={WebConf},
  year={2018}
}

@inproceedings{tsioutsiouliklis2022link,
  title={Link recommendations for PageRank fairness},
  author={Tsioutsiouliklis, Sotiris and Pitoura, Evaggelia and Semertzidis, Konstantinos and Tsaparas, Panayiotis},
  booktitle={WebConf},
  year={2022}
}

@inproceedings{tsioutsiouliklis2021fairness,
  title={Fairness-aware pagerank},
  author={Tsioutsiouliklis, Sotiris and Pitoura, Evaggelia and Tsaparas, Panayiotis and Kleftakis, Ilias and Mamoulis, Nikos},
  booktitle={WebConf},
  year={2021}
}

@article{karimi2018homophily,
  title={Homophily influences ranking of minorities in social networks},
  author={Karimi, Fariba and G{\'e}nois, Mathieu and Wagner, Claudia and Singer, Philipp and Strohmaier, Markus},
  journal={Scientific reports},
  volume={8},
  number={1},
  pages={11077},
  year={2018},
  publisher={Nature Publishing Group UK London}
}

@inproceedings{celis2018ranking,
  title={Ranking with Fairness Constraints},
  author={Celis, L Elisa and Straszak, Damian and Vishnoi, Nisheeth K},
  booktitle={ICALP},
  year={2018},
}

@inproceedings{xu2018powerful,
  title={How powerful are graph neural networks?},
  author={Xu, Keyulu and Hu, Weihua and Leskovec, Jure and Jegelka, Stefanie},
  booktitle={ICLR},
  year={2019}
}

@inproceedings{mattos2025attribute,
  author    = {João Mattos and Zexi Huang and Mert Kosan and Ambuj Singh and Arlei Silva},
  title     = {Attribute-Enhanced Similarity Ranking for Sparse Link Prediction},
  booktitle = {SIGKDD},
  year      = {2025},
}

@article{mcdonald2009networks,
  title={Networks of opportunity: Gender, race, and job leads},
  author={McDonald, Steve and Lin, Nan and Ao, Dan},
  journal={Social Problems},
  volume={56},
  number={3},
  pages={385--402},
  year={2009},
  publisher={Oxford University Press Oxford, UK}
}

@article{calvo2004effects,
  title={The effects of social networks on employment and inequality},
  author={Calvo-Armengol, Antoni and Jackson, Matthew O},
  journal={American economic review},
  volume={94},
  number={3},
  pages={426--454},
  year={2004},
  publisher={American Economic Association}
}

@inproceedings{li_dyadic_2021,
  title={On dyadic fairness: Exploring and mitigating bias in graph connections},
  author={Li, Peizhao and Wang, Yifei and Zhao, Han and Hong, Pengyu and Liu, Hongfu},
  booktitle={ICLR},
  year={2021}
}

@article{hofstra2017sources,
  title={Sources of segregation in social networks: A novel approach using Facebook},
  author={Hofstra, Bas and Corten, Rense and Van Tubergen, Frank and Ellison, Nicole B},
  journal={American Sociological Review},
  volume={82},
  number={3},
  pages={625--656},
  year={2017},
  publisher={SAGE Publications Sage CA: Los Angeles, CA}
}

@inproceedings{gupta2021correcting,
  title={Correcting exposure bias for link recommendation},
  author={Gupta, Shantanu and Wang, Hao and Lipton, Zachary and Wang, Yuyang},
  booktitle={ICML},
  year={2021},
}

@inproceedings{xian2023fair,
  title={Fair and optimal classification via post-processing},
  author={Xian, Ruicheng and Yin, Lang and Zhao, Han},
  booktitle={ICML},
  year={2023},
}

@inproceedings{tifrea2023frappe,
  title={Frapp{\'e}: A group fairness framework for post-processing everything},
  author={Tifrea, Alexandru and Lahoti, Preethi and Packer, Ben and Halpern, Yoni and Beirami, Ahmad and Prost, Flavien},
  booktitle={ICML},
  year={2024}
}

@inproceedings{Fey/Lenssen/2019,
  title={Fast Graph Representation Learning with {PyTorch Geometric}},
  author={Fey, Matthias and Lenssen, Jan E.},
  booktitle={ICLR RLGM},
  year={2019},
}

@article{dong2023fairness,
  title={Fairness in graph mining: A survey},
  author={Dong, Yushun and Ma, Jing and Wang, Song and Chen, Chen and Li, Jundong},
  journal={TKDE},
  year={2023},
  publisher={IEEE}
}

@inproceedings{edwards2015censoring,
  title={Censoring representations with an adversary},
  author={Edwards, Harrison and Storkey, Amos},
  booktitle={ICLR},
  year={2016}
}

@inproceedings{li_fairlp_2022,
	title = {{FairLP}: {Towards} {Fair} {Link} {Prediction} on {Social} {Network} {Graphs}},
	author = {Li, Yanying and Wang, Xiuling and Ning, Yue and Wang, Hui},
    booktitle={ICWSM},
	year = {2022},
}

@inproceedings{current_fairegm_2022,
	title = {{FairEGM}: {Fair} {Link} {Prediction} and {Recommendation} via {Emulated} {Graph} {Modification}},
	booktitle = {EAMO},
	author = {Current, Sean and He, Yuntian and Gurukar, Saket and Parthasarathy, Srinivasan},
	year = {2022},
}

@inproceedings{masrour_bursting_2020,
	title = {Bursting the {Filter} {Bubble}: {Fairness}-{Aware} {Network} {Link} {Prediction}},
	booktitle = {AAAI},
	author = {Masrour, Farzan and Wilson, Tyler and Yan, Heng and Tan, Pang-Ning and Esfahanian, Abdol},
	year = {2020},
}

@inproceedings{dong_edits_2022,
	title = {{EDITS}: {Modeling} and {Mitigating} {Data} {Bias} for {Graph} {Neural} {Networks}},
	booktitle = {WebConf},
	author = {Dong, Yushun and Liu, Ninghao and Jalaian, Brian and Li, Jundong},
	year = {2022},
}

@inproceedings{ling_learning_2023,
	title = {{Learning} {Fair} {Graph} {Representations} {via} {Automated} {Data} {Augmentations}},
	author = {Ling, Hongyi and Jiang, Zhimeng and Luo, Youzhi and Ji, Shuiwang and Zou, Na},
	year = {2023},
    booktitle={ICLR},
}

@inproceedings{dong_reliant_2023,
	title = {{RELIANT}: {Fair} {Knowledge} {Distillation} for {Graph} {Neural} {Networks}},
	author = {Dong, Yushun and Zhang, Binchi and Yuan, Yiling and Zou, Na and Wang, Qi and Li, Jundong},
	year = {2023},
	booktitle={SDM}
}

@inproceedings{agarwal_towards_2021,
	title = {Towards a {Unified} {Framework} for {Fair} and {Stable} {Graph} {Representation} {Learning}},
	author = {Agarwal, Chirag and Lakkaraju, Himabindu and Zitnik, Marinka},
	year = {2021},
	booktitle={UAI}
}

@inproceedings{dai_say_2021,
	title = {Say {No} to the {Discrimination}: {Learning} {Fair} {Graph} {Neural} {Networks} with {Limited} {Sensitive} {Attribute} {Information}},
	year = {2021},
	booktitle={WSDM}
}

@inproceedings{kusner_counterfactual_2017,
	title = {Counterfactual {Fairness}},
	booktitle = {NeurIPS},
	author = {Kusner, Matt J and Loftus, Joshua and Russell, Chris and Silva, Ricardo},
	year = {2017},
}

@inproceedings{wang_improving_2022,
  title={Improving fairness in graph neural networks via mitigating sensitive attribute leakage},
  author={Wang, Yu and Zhao, Yuying and Dong, Yushun and Chen, Huiyuan and Li, Jundong and Derr, Tyler},
  booktitle={SIGKDD},
  year={2022}
}

@inproceedings{zhu_one_2024,
  title={One fits all: Learning fair graph neural networks for various sensitive attributes},
  author={Zhu, Yuchang and Li, Jintang and Bian, Yatao and Zheng, Zibin and Chen, Liang},
  booktitle={SIGKDD},
  year={2024}
}

@inproceedings{zhu_devil_2024,
  title={The devil is in the data: Learning fair graph neural networks via partial knowledge distillation},
  author={Zhu, Yuchang and Li, Jintang and Chen, Liang and Zheng, Zibin},
  booktitle={WSDM},
  year={2024}
}

@inproceedings{zehlike_fair_2017,
	title = {{FA}*{IR}: {A} {Fair} {Top}-k {Ranking} {Algorithm}},
	booktitle = {CIKM},
	author = {Zehlike, Meike and Bonchi, Francesco and Castillo, Carlos and Hajian, Sara and Megahed, Mohamed and Baeza-Yates, Ricardo},
	year = {2017},
}

@inproceedings{geyik_fairness-aware_2019,
	title = {Fairness-{Aware} {Ranking} in {Search} \& {Recommendation} {Systems} with {Application} to {LinkedIn} {Talent} {Search}},
	booktitle = {SIGKDD},
	author = {Geyik, Sahin Cem and Ambler, Stuart and Kenthapadi, Krishnaram},
	year = {2019},
}

@inproceedings{gorantla_problem_2021,
	title = {On the {Problem} of {Underranking} in {Group}-{Fair} {Ranking}},
	booktitle = {ICML},
	author = {Gorantla, Sruthi and Deshpande, Amit and Louis, Anand},
	year = {2021}
}

@inproceedings{singh_policy_2019,
  title={Policy learning for fairness in ranking},
  author={Singh, Ashudeep and Joachims, Thorsten},
  booktitle={NeurIPS},
  year={2019}
}

@inproceedings{zehlike_reducing_2020,
	title = {Reducing {Disparate} {Exposure} in {Ranking}: {A} {Learning} {To} {Rank} {Approach}},
	booktitle = {WebConf},
	author = {Zehlike, Meike and Castillo, Carlos},
	year = {2020},
}

@inproceedings{singh_fairness_2018,
	title = {Fairness of {Exposure} in {Rankings}},
	booktitle = {SIGKDD},
	author = {Singh, Ashudeep and Joachims, Thorsten},
	year = {2018},
}

@inproceedings{dwork_decoupled_2018,
	title = {Decoupled {Classifiers} for {Group}-{Fair} and {Efficient} {Machine} {Learning}},
	booktitle = {FAccT},
	author = {Dwork, Cynthia and Immorlica, Nicole and Kalai, Adam Tauman and Leiserson, Max},
	year = {2018},
}

@inproceedings{dwork_fairness_2011,
  title={Fairness through awareness},
  author={Dwork, Cynthia and Hardt, Moritz and Pitassi, Toniann and Reingold, Omer and Zemel, Richard},
  booktitle={ITCS},
  year={2012}
}

@inproceedings{joachims_accurately_nodate,
  title={Accurately interpreting clickthrough data as implicit feedback},
  author={Joachims, Thorsten and Granka, Laura and Pan, Bing and Hembrooke, Helene and Gay, Geri},
  booktitle={SIGIR},
  year={2005}
}

@inproceedings{rahman_fairwalk_2019,
    title = {Fairwalk: {Towards} {Fair} {Graph} {Embedding}},
    booktitle = {IJCAI},
    author = {Rahman, Tahleen and Surma, Bartlomiej and Backes, Michael and Zhang, Yang},
    year = {2019},
}

@inproceedings{kotary_end--end_2021,
  title={End-to-end learning for fair ranking systems},
  author={Kotary, James and Fioretto, Ferdinando and Van Hentenryck, Pascal and Zhu, Ziwei},
  booktitle={WebConf},
  year={2022}
}

@inproceedings{yadav_policy-gradient_2021,
    title = {Policy-{Gradient} {Training} of {Fair} and {Unbiased} {Ranking} {Functions}},
    booktitle = {SIGIR},
    author = {Yadav, Himank and Du, Zhengxiao and Joachims, Thorsten},
    year = {2021},
}

@inproceedings{zhang_link_2018,
    title = {Link {Prediction} {Based} on {Graph} {Neural} {Networks}},
    booktitle = {NeurIPS},
    author = {Zhang, Muhan and Chen, Yixin},
    year = {2018},
}

@article{liben2007link,
  title={The link-prediction problem for social networks},
  author={Liben-Nowell, David and Kleinberg, Jon},
  journal={Journal of the American society for information science and technology},
  volume={58},
  number={7},
  pages={1019--1031},
  year={2007},
  publisher={Wiley Online Library}
}

@inproceedings{pan2021neural,
  title={Neural Link Prediction with Walk Pooling},
  author={Pan, Liming and Shi, Cheng and Dokmani{\'c}, Ivan},
  booktitle={ICLR},
  year={2022}
}

@inproceedings{zhu2021neural,
  title={Neural bellman-ford networks: A general graph neural network framework for link prediction},
  author={Zhu, Zhaocheng and Zhang, Zuobai and Xhonneux, Louis-Pascal and Tang, Jian},
  booktitle={NeurIPS},
  year={2021}
}

@inproceedings{chamberlain2022graph,
  title={Graph Neural Networks for Link Prediction with Subgraph Sketching},
  author={Chamberlain, Benjamin Paul and Shirobokov, Sergey and Rossi, Emanuele and Frasca, Fabrizio and Markovich, Thomas and Hammerla, Nils and Bronstein, Michael M and Hansmire, Max},
  booktitle={ICLR},
  year={2023}
}

@article{hanretiring,
  title={Retiring {$\Delta DP$}: New Distribution-Level Metrics for Demographic Parity},
  author={Han, Xiaotian and Jiang, Zhimeng and Jin, Hongye and Liu, Zirui and Zou, Na and Wang, Qifan and Hu, Xia},
  journal={TMLR},
  year={2023}
}

@article{draws2021assessing,
  title={Assessing viewpoint diversity in search results using ranking fairness metrics},
  author={Draws, Tim and Tintarev, Nava and Gadiraju, Ujwal},
  journal={SIGKDD Explorations},
  volume={23},
  number={1},
  pages={50--58},
  year={2021},
  publisher={ACM New York, NY, USA}
}

@inproceedings{wang_neural_2024,
  title={Neural Common Neighbor with Completion for Link Prediction},
  author={Wang, Xiyuan and Yang, Haotong and Zhang, Muhan},
  booktitle={ICLR},
   year={2024}
}

@article{cinelli2021echo,
  title={The echo chamber effect on social media},
  author={Cinelli, Matteo and De Francisci Morales, Gianmarco and Galeazzi, Alessandro and Quattrociocchi, Walter and Starnini, Michele},
  journal={PNAS},
  volume={118},
  number={9},
  pages={e2023301118},
  year={2021},
  publisher={National Academy of Sciences}
}

@article{bakshy2015exposure,
  title={Exposure to ideologically diverse news and opinion on Facebook},
  author={Bakshy, Eytan and Messing, Solomon and Adamic, Lada A},
  journal={Science},
  volume={348},
  number={6239},
  pages={1130--1132},
  year={2015},
  publisher={American Association for the Advancement of Science}
}

@inproceedings{kleinberg2024calibrated,
  title={Calibrated recommendations for users with decaying attention},
  author={Kleinberg, Jon and Ryu, Emily and Tardos, {\'E}va},
  booktitle={SAGT},
  year={2024},
}

@inproceedings{tsioutsiouliklis_fairness-aware_2021,
  title={Fairness-aware pagerank},
  author={Tsioutsiouliklis, Sotiris and Pitoura, Evaggelia and Tsaparas, Panayiotis and Kleftakis, Ilias and Mamoulis, Nikos},
  booktitle={WebConf},
  year={2021}
}

@inproceedings{kearns2018preventing,
  title={Preventing fairness gerrymandering: Auditing and learning for subgroup fairness},
  author={Kearns, Michael and Neel, Seth and Roth, Aaron and Wu, Zhiwei Steven},
  booktitle={ICML},
  year={2018},
}

\appendix

\newpage

\section{Example - Limited Expressive Power Affects Fair Link Prediction}
\label{ap::limited_expressive_power}

\begin{figure}[H]
    \centering \includegraphics[width=0.3\linewidth]{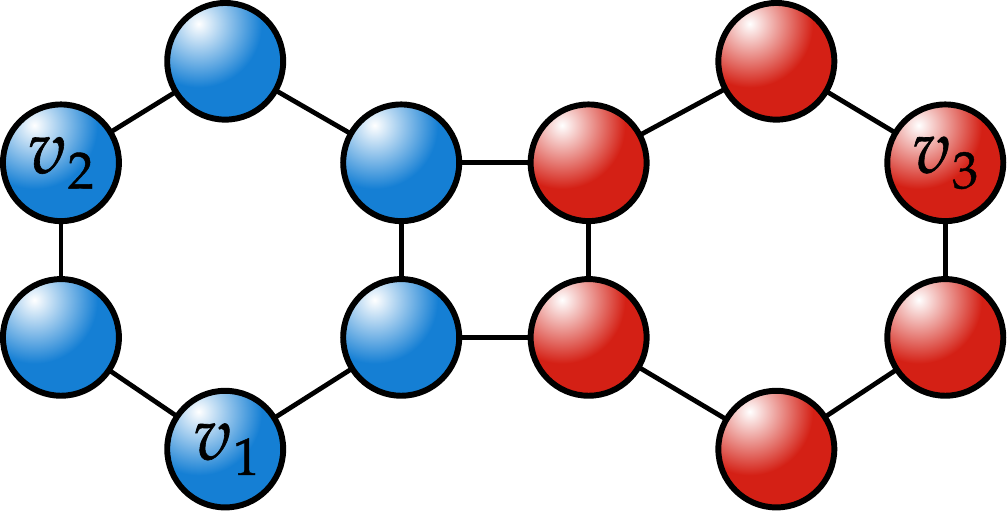}
    \caption{1-WL GNN Indistinguishability in Pair Representations.This graph illustrates a structural limitation of 1-WL GNNs in distinguishing node pairs for fairness-aware link prediction. Nodes are colored by sensitive group (blue and red), with labeled nodes $v_1$, $v_2$, and $v_3$. Despite $v_2$ and$v_3$ belonging to different groups, both are structurally symmetric with respect to $v_1$. As a result, standard message-passing GNNs produce nearly identical embeddings for $(v_1, v_2)$ and $(v_1, v_3)$, failing to capture the demographic asymmetry between these node pairs. This highlights a key expressivity limitation of 1-WL GNNs for fair link prediction.
}
    \label{fig:enter-label}
\end{figure}

\section{Experimental Results for NDKL Bounds}
\label{ap::ndkl_experimental_bounds}

\begin{figure}[h]
    \includegraphics[width=1.\linewidth]{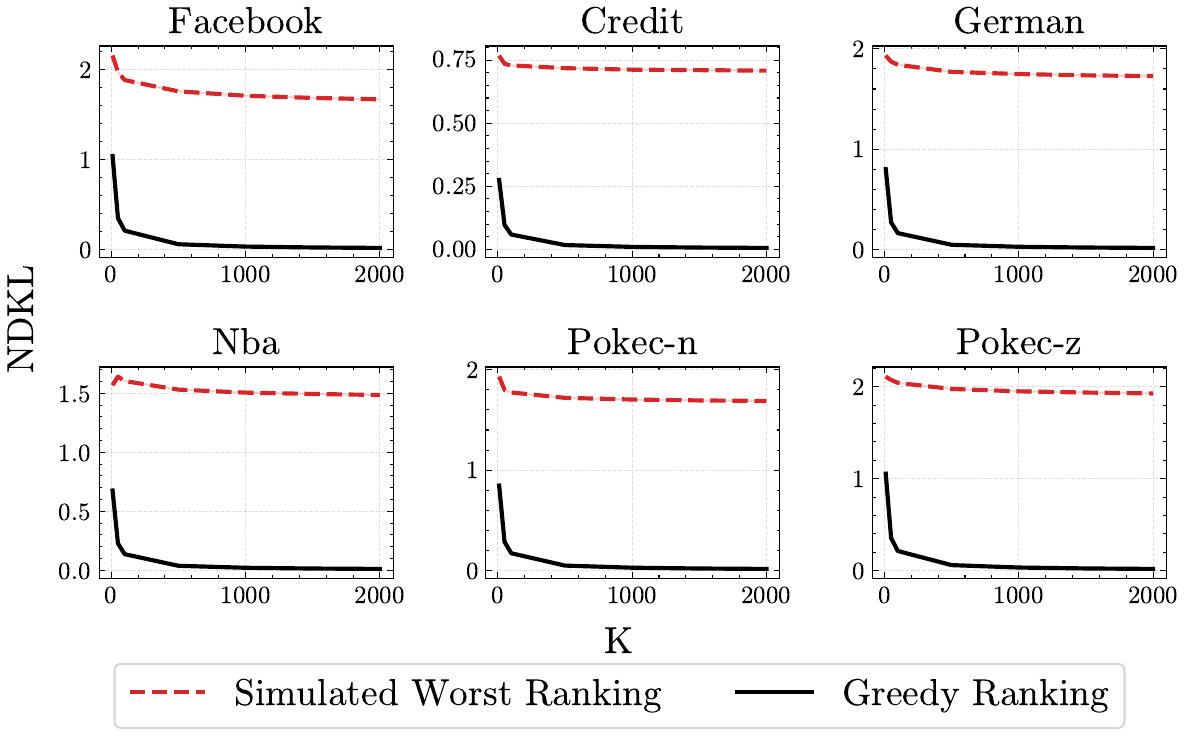}
    \caption{NDKL gap for optimal \dempar for all six datasets, varying $k$. The \textcolor{red}{\textbf{red}} curve denotes the worst ranking obtained with the optimal proportions of each pair type, while the \textbf{black} curve denotes the ranking outputted by the Algorithm \ref{alg:greedy_dkl}. Despite both curves obtaining the same (optimal) value of \dempar, there is a very significant gap between the NDKL measurements.}
    \label{fig:gap_experiment}
\end{figure}

We illustrate the NDKL gap for optimal \dempar in Figure \ref{fig:gap_experiment}. In all datasets, both the Worst Ranking and the Greedy Ranking obtain the same values of \dempar, while having drastic differences in NDKL.

\newpage

\section{Example -  Optimal \dempar}
\label{ap:toy_example}

We achieve optimal \dempar by minimizing $\min_x|\frac{x}{|E_{\text{intra}}|} - \frac{x-k}{|E_{\text{inter}}|}|$, in which $x$ is the number of non-protected pairs ($E_{\text{intra}}$) to be recommended and $k$ is the total number of pairs to be predicted.

In the example, we consider a top-$10$ ranking ($k=10$), and considering the connectivity of the graph given as an example, we $|E_{\text{intra}}|=3$ and $|E_{\text{inter}}|=11$, obtaining $\min_x|\frac{x}{13} - \frac{x-10}{11}|\approx0.06$, with $x=2$ as the optimal value. 

Figure \ref{fig:placeholder1} demonstrates two possible scenarios of recommendations based on the obtained values of $x$ and \dempar. The scenario on the left overrepresents $\ENSNS$ pairs (blue), whereas the scenario on the right provides more balanced recommendations considering the community structure of the original graph (edges in black). Both scenarios obtain the optimal value of \dempar due to the dyadic assumption of the metric.

\section{Proof of Theorem \ref{theorem::ndkl}}

\begin{proof}

The lower bound is attained when every top-$k$ prefix distribution exactly matches $\boldsymbol{\pi}$, i.e., $\hat{\boldsymbol{\pi}}_k = \boldsymbol{\pi}$ for all $k$. This corresponds to a perfectly interleaved ranking aligned with the target group proportions. The upper bound is attained when the rarest group (i.e., the one with $\pi_i = \min_j \pi_j$) dominates the top of the ranking, followed by the others, creating maximal prefix divergence while preserving demographic parity globally.

These bounds illustrate that NDKL can distinguish among rankings that are equally fair under demographic parity by accounting for the fairness of ranking prefixes.
    
\end{proof}

\section{NDKL Bounds for Arbitrary Sensitive Group Distributions}

We generalize the NDKL bounds to the case where the target distribution $\boldsymbol{\pi} = [\pi_1, \pi_2, \dots, \pi_G]$ is defined over $G \geq 2$ sensitive groups, with $\sum_{i=1}^{G} \pi_i = 1$.

\paragraph{NDKL Bounds.} 
Under the demographic parity constraint, the NDKL score satisfies:
\begin{equation}
    0 \;\leq\; \mathrm{NDKL} \;\leq\; \max_{i \in \{1, \dots, G\}} \log \frac{1}{\pi_i}.
\end{equation}

\paragraph{Lower Bound (Zero).}
The minimum NDKL is achieved when every prefix distribution exactly matches the target:
\[
\hat{\boldsymbol{\pi}}_k = \boldsymbol{\pi} \quad \forall k \in \{1, \dots, N\}.
\]
In this case, each KL term vanishes, and $\mathrm{NDKL} = 0$.

\paragraph{Upper Bound.}
To maximize NDKL while maintaining demographic parity globally, the worst-case ranking front-loads elements from the group with the smallest mass in $\boldsymbol{\pi}$, say $i^\ast = \arg\min_i \pi_i$. Then for early $k$, the prefix $\hat{\boldsymbol{\pi}}_k \approx \mathbf{e}_{i^\ast}$, yielding
\[
D_{\mathrm{KL}}(\hat{\boldsymbol{\pi}}_k \,\|\, \boldsymbol{\pi}) \approx \log \frac{1}{\pi_{i^\ast}} = \max_i \log \frac{1}{\pi_i}.
\]
As $\log_2(k+1)$ weights decrease with $k$, early skewed prefixes dominate the NDKL sum. Therefore,
\[
\mathrm{NDKL} \leq \frac{1}{Z} \sum_{k=1}^{N} \frac{1}{\log_2(k+1)} \cdot \max_{i} \log \frac{1}{\pi_i} = \max_{i} \log \frac{1}{\pi_i}.
\]

\paragraph{Conclusion.}
The NDKL metric provides a continuum of fairness measurement within the space of rankings that all satisfy demographic parity globally. While the lower bound corresponds to a perfectly interleaved fair ranking, the upper bound captures the worst-case prefix bias permissible under demographic parity.

\section{MORAL - Algorithm pseudocode}
\label{ap::moral_pseudocode}

\begin{algorithm}[H]
\caption{MORAL: Multi-Output Ranking Aggregation for Link Fairness}
\label{alg:greedy_dkl}
\KwIn{
    \begin{itemize}
        \item Candidate sets $\mathcal{C}_j = \{(u,v, \text{score})\}$ for each group $j \in \{0,1,2\}$ (sorted by descending score);
        \item Target distribution $\boldsymbol{\pi} = (\pi_0, \pi_1, \pi_2)$;
        \item Total output size $n$.
    \end{itemize}
}
\KwOut{Ranking list $\mathbf{R}$ of $n$ predicted edges with assigned group labels}

Initialize exposure counts: $\boldsymbol{c} \leftarrow (0, 0, 0)$\;
Initialize output ranking: $\mathbf{R} \leftarrow [\;]$\;

\For{$t \leftarrow 1$ \KwTo $n$}{
    Initialize best objective: $\text{min\_kl} \leftarrow \infty$, $\text{selected\_group} \leftarrow -1$, $\text{selected\_edge} \leftarrow \text{None}$\;

    \ForEach{group $j \in \{0,1,2\}$ such that $\mathcal{C}_j$ is not empty}{
        Let $(u,v, \text{score}) \leftarrow$ top element in $\mathcal{C}_j$\;
        
        Temporarily update counts: $c_j' \leftarrow c_j + 1$, $q_j' \leftarrow \frac{c_j'}{t}$, $q_{j'\neq j} \leftarrow \frac{c_{j'}}{t}$\;
        
        Compute KL divergence: $D_{\mathrm{KL}}( \mathbf{q}' \| \boldsymbol{\pi} )$\;
        
        \If{this KL is lower than $\text{min\_kl}$}{
            Update $\text{min\_kl} \leftarrow D_{\mathrm{KL}}$, $\text{selected\_group} \leftarrow j$, $\text{selected\_edge} \leftarrow (u,v)$\;
        }
    }

    Append $(\text{selected\_edge}, \text{selected\_group})$ to $\mathbf{R}$\;
    
    Remove top element from $\mathcal{C}_{\text{selected\_group}}$\;
    
    Update $c_{\text{selected\_group}} \leftarrow c_{\text{selected\_group}} + 1$\;
}

\Return{$\mathbf{R}$}
\end{algorithm}






\section{Additional Performance Results}
\label{ap::additional_performance_results}

\begin{table}
\caption{Performance comparison on the Cora dataset using NDKL$\downarrow$ and $prec@1000$$\uparrow$. Best (lowest) NDKL is \textbf{bold}. Best (highest) $prec@1000$ is \underline{underlined}.}

\begin{tabular}{llc}
\toprule
 \textbf{Method} &  & \textbf{Cora} \\
\midrule
\multirow{2}{*}{NIFTY GAE} & NDKL & 0.0540 $\pm$ 0.0030 \\
 & $prec@1000$ & \underline{0.7300 $\pm$ 0.2970} \\
\midrule
\multirow{2}{*}{FairWalk} & NDKL & 0.0522 $\pm$ 0.0002 \\
 & $prec@1000$ & 0.5110 $\pm$ 0.0030 \\
\midrule
\multirow{2}{*}{DELTR GAE} & NDKL & 0.0484 $\pm$ 0.0001 \\
 & $prec@1000$ & 0.5225 $\pm$ 0.0007 \\
\midrule
\multirow{2}{*}{MORAL} & NDKL & \textbf{0.0471 $\pm$ 0.0000} \\
 & $prec@1000$ & 0.7229 $\pm$ 0.0010 \\
\midrule
\bottomrule
\end{tabular}
\label{tab:cora_results}
\end{table}

In addition to the $prec@1000$ and NDKL results exposed in Subsection \ref{subsec::baselines_ranking_comparison}, we conduct the evaluation on other performance ($prec@100$, $NDCG$, $AP$) and fairness ($\Delta_{DP}$) exposed in \cref{tab:results_prec_100,tab:ap_ndkl,tab:hits_ndkl,tab:ndcg_ndkl,tab:parity_precision}, respectively. We highlight that even though GraphAIR does not obtain the lowest $NDKL$ values, it is the overall most fair outputs considering $\Delta_{DP}$, demonstrating the limitation of the metric. 

We simulate a multi-sensitive attribute scenario by considering a multi-label network, i.e. Cora, and label as a sensitive attribute value. In order to avoid the combinatorial cost of training $k\choose2$ classifiers, we train a single model per sensitive attribute value in a one vs. all classification fashion. We expose the results on a simulated multi-sensitive attribute scenario conducted on Cora on Table \ref{tab:cora_results}.

\begin{table*}[htb]
\centering
\small
\setlength{\tabcolsep}{4pt}
\caption{Fairness performance comparison of all approaches considered ($k=100$). Lower \textit{NDKL} and higher \textit{prec@100} are better. Best NDKL and \textit{prec@100} values are in \textbf{bold} and \underline{underline}, respectively.}
\begin{tabular}{llcccccc}
\toprule
\textbf{Method} &  & \textbf{Facebook} & \textbf{Credit} & \textbf{German} & \textbf{NBA} & \textbf{Pokec-n} & \textbf{Pokec-z} \\
\midrule
\multirow{2}{*}{UGE} & NDKL & 0.29 $\pm$ 0.01 & 0.88 $\pm$ 0.00 & 0.43 $\pm$ 0.18 & 0.38 $\pm$ 0.11 & 0.34 $\pm$ 0.03 & 0.35 $\pm$ 0.02 \\
 & $prec@100$ & \underline{0.99 $\pm$ 0.02} & \underline{1.00 $\pm$ 0.00} & 0.81 $\pm$ 0.07 & 0.62 $\pm$ 0.02 & 0.96 $\pm$ 0.04 & 0.97 $\pm$ 0.03 \\

\midrule

\multirow{2}{*}{EDITS} & NDKL & 1.05 $\pm$ 0.37 & 0.22 $\pm$ 0.13 & 0.37 $\pm$ 0.07 & 0.44 $\pm$ 0.10 & OOM & OOM \\
 & $prec@100$ & \underline{0.99 $\pm$ 0.01} & 1.00 $\pm$ 0.00 & 1.00 $\pm$ 0.00 & 0.48 $\pm$ 0.08 & OOM & OOM \\
 
\midrule
\multirow{2}{*}{GRAPHAIR} & NDKL & 0.74 $\pm$ 0.16 & 0.88 $\pm$ 0.00 & 0.38 $\pm$ 0.11 & 0.50 $\pm$ 0.03 & 0.50 $\pm$ 0.17 & 0.97 $\pm$ 0.70 \\
 & $prec@100$ & 0.98 $\pm$ 0.01 & 1.00 $\pm$ 0.01 & 0.84 $\pm$ 0.00 & 0.84 $\pm$ 0.01 & 0.98 $\pm$ 0.01 & 1.00 $\pm$ 0.00 \\
 
\midrule
\multirow{2}{*}{FairWalk} & NDKL & 0.35 $\pm$ 0.04 & 0.36 $\pm$ 0.16 & 0.55 $\pm$ 0.15 & 0.37 $\pm$ 0.08 & 0.35 $\pm$ 0.08 & 0.40 $\pm$ 0.02 \\
 & $prec@100$ & 0.98 $\pm$ 0.01 & 1.00 $\pm$ 0.00 & 0.97 $\pm$ 0.01 & 0.60 $\pm$ 0.05 & 1.00 $\pm$ 0.00 & 1.00 $\pm$ 0.00 \\
 
\midrule
\multirow{2}{*}{FairAdj} & NDKL & 0.56 $\pm$ 0.29 & OOM & 0.49 $\pm$ 0.05 & 0.60 $\pm$ 0.30 & OOM & OOM \\
 & $prec@100$ & 0.30 $\pm$ 0.06 & OOM & 0.40 $\pm$ 0.02 & 0.68 $\pm$ 0.02 & OOM & OOM \\
 
\midrule
\multirow{2}{*}{DetConstSort} & NDKL & 0.90 $\pm$ 0.00 & 0.28 $\pm$ 0.00 & 0.26 $\pm$ 0.00 & 0.53 $\pm$ 0.00 & 0.35 $\pm$ 0.00 & 1.05 $\pm$ 0.00 \\
 & $prec@100$ & 0.01 $\pm$ 0.00 & 0.00 $\pm$ 0.00 & 0.58 $\pm$ 0.00 & 0.27 $\pm$ 0.00 & 0.07 $\pm$ 0.00 & 0.01 $\pm$ 0.00 \\
\midrule
\multirow{2}{*}{DELTR} & NDKL & 0.59 $\pm$ 0.18 & 0.18 $\pm$ 0.03 & 0.43 $\pm$ 0.21 & 0.47 $\pm$ 0.02 & 1.02 $\pm$ 1.17 & 0.90 $\pm$ 0.84 \\
 & $prec@100$ & 0.94 $\pm$ 0.08 & \underline{1.00 $\pm$ 0.00} & \underline{1.00 $\pm$ 0.00} & 0.39 $\pm$ 0.45 & 0.76 $\pm$ 0.13 & 0.72 $\pm$ 0.12 \\
\midrule
\multirow{2}{*}{MORAL} & NDKL & \textbf{0.22 $\pm$ 0.00} & \textbf{0.06 $\pm$ 0.00} & \textbf{0.17 $\pm$ 0.00} & \textbf{0.14 $\pm$ 0.00} & \textbf{0.18 $\pm$ 0.00} & \textbf{0.23 $\pm$ 0.00} \\
 & $prec@100$ & 0.96 $\pm$ 0.00 & \underline{1.00 $\pm$ 0.00} & 0.99 $\pm$ 0.00 & \underline{0.87 $\pm$ 0.04} & 0.98 $\pm$ 0.00 & 0.99 $\pm$ 0.01 \\
\bottomrule
\end{tabular}
\label{tab:results_prec_100}
\end{table*}

\begin{table*}
\centering
\small
\setlength{\tabcolsep}{4pt}
\caption{Performance comparison (NDKL↓ and AP↑). Best AP per dataset is \underline{underlined}. Best NDKL per dataset is \textbf{bold}. OOM = method ran out-of-memory.}
\begin{tabular}{llcccccc}
\toprule
\textbf{Method} &  & \textbf{Facebook} & \textbf{Credit} & \textbf{German} & \textbf{NBA} & \textbf{Pokec-n} & \textbf{Pokec-z} \\
\midrule

\multirow{2}{*}{UGE}
& NDKL
& 0.05 $\pm$ 0.00
& 0.80 $\pm$ 0.07
& 0.08 $\pm$ 0.03
& 0.07 $\pm$ 0.02
& 0.06 $\pm$ 0.00
& 0.06 $\pm$ 0.00 \\
& AP
& 0.93 $\pm$ 0.00
& 0.85 $\pm$ 0.01
& 0.60 $\pm$ 0.01
& 0.56 $\pm$ 0.01
& 0.63 $\pm$ 0.03
& 0.63 $\pm$ 0.02 \\
\midrule

\multirow{2}{*}{EDITS GAE}
& NDKL
& 0.21 $\pm$ 0.07
& 0.04 $\pm$ 0.02
& 0.08 $\pm$ 0.02
& 0.08 $\pm$ 0.01
& OOM & OOM \\
& AP
& 0.78 $\pm$ 0.01
& 0.63 $\pm$ 0.01
& 0.50 $\pm$ 0.00
& 0.50 $\pm$ 0.00
& OOM & OOM \\
\midrule

\multirow{2}{*}{GRAPHAIR}
& NDKL
& 0.13 $\pm$ 0.03
& 0.67 $\pm$ 0.22
& 0.07 $\pm$ 0.02
& 0.09 $\pm$ 0.01
& 0.09 $\pm$ 0.03
& 0.26 $\pm$ 0.25 \\
& AP
& 0.90 $\pm$ 0.00
& 0.90 $\pm$ 0.02
& 0.62 $\pm$ 0.00
& 0.65 $\pm$ 0.01
& 0.75 $\pm$ 0.05
& 0.66 $\pm$ 0.23 \\
\midrule

\multirow{2}{*}{FairEGM}
& NDKL
& 0.09 $\pm$ 0.01
& 0.11 $\pm$ 0.00
& 0.05 $\pm$ 0.01
& 0.07 $\pm$ 0.01
& OOM & OOM \\
& AP
& \underline{0.94 $\pm$ 0.00}
& 0.96 $\pm$ 0.00
& 0.56 $\pm$ 0.00
& 0.58 $\pm$ 0.01
& OOM & OOM \\
\midrule

\multirow{2}{*}{FairLP}
& NDKL
& 0.18 $\pm$ 0.00
& OOM
& 0.06 $\pm$ 0.00
& 0.20 $\pm$ 0.00
& OOM & OOM \\
& AP
& \underline{0.95 $\pm$ 0.00}
& OOM
& 0.84 $\pm$ 0.00
& 0.78 $\pm$ 0.00
& OOM & OOM \\
\midrule

\multirow{2}{*}{FairWalk}
& NDKL
& 0.06 $\pm$ 0.01
& 0.06 $\pm$ 0.03
& 0.11 $\pm$ 0.02
& 0.06 $\pm$ 0.01
& 0.07 $\pm$ 0.01
& 0.07 $\pm$ 0.00 \\
& AP
& 0.91 $\pm$ 0.00
& 0.98 $\pm$ 0.00
& 0.86 $\pm$ 0.00
& 0.54 $\pm$ 0.01
& 0.86 $\pm$ 0.00
& 0.86 $\pm$ 0.00 \\
\midrule

\multirow{2}{*}{FairAdj}
& NDKL
& 0.10 $\pm$ 0.05
& OOM
& 0.10 $\pm$ 0.01
& 0.11 $\pm$ 0.05
& OOM & OOM \\
& AP
& 0.55 $\pm$ 0.01
& OOM
& 0.51 $\pm$ 0.00
& 0.53 $\pm$ 0.01
& OOM & OOM \\
\midrule

\multirow{2}{*}{DetConstSort GAE}
& NDKL
& 0.15 $\pm$ 0.00
& 0.06 $\pm$ 0.00
& 0.04 $\pm$ 0.00
& 0.09 $\pm$ 0.00
& 0.07 $\pm$ 0.00
& 0.23 $\pm$ 0.00 \\
& AP
& 0.05 $\pm$ 0.00
& 0.00 $\pm$ 0.00
& 0.56 $\pm$ 0.00
& 0.28 $\pm$ 0.00
& 0.07 $\pm$ 0.00
& 0.02 $\pm$ 0.00 \\
\midrule

\multirow{2}{*}{DELTR GAE}
& NDKL
& 0.10 $\pm$ 0.03
& 0.03 $\pm$ 0.00
& 0.09 $\pm$ 0.06
& 0.09 $\pm$ 0.02
& 0.23 $\pm$ 0.23
& 0.22 $\pm$ 0.20 \\
& AP
& 0.78 $\pm$ 0.04
& 0.60 $\pm$ 0.02
& 0.50 $\pm$ 0.03
& 0.47 $\pm$ 0.14
& 0.44 $\pm$ 0.08
& 0.41 $\pm$ 0.06 \\
\midrule

\multirow{2}{*}{ThreeClass GAE}
& NDKL
& \textbf{0.04 $\pm$ 0.00}
& \textbf{0.01 $\pm$ 0.00}
& \textbf{0.03 $\pm$ 0.00}
& \textbf{0.03 $\pm$ 0.00}
& \textbf{0.02 $\pm$ 0.00}
& \textbf{0.03 $\pm$ 0.00} \\
& AP
& \underline{0.94 $\pm$ 0.00}
& \underline{1.00 $\pm$ 0.00}
& \underline{0.96 $\pm$ 0.01}
& \underline{0.80 $\pm$ 0.00}
& \underline{0.98 $\pm$ 0.00}
& \underline{0.98 $\pm$ 0.00} \\
\bottomrule

\end{tabular}
\label{tab:ap_ndkl}
\end{table*}

\begin{table*}
\centering
\small
\setlength{\tabcolsep}{4pt}
\caption{Performance comparison ($hits@1000$↑ and NDKL↓). Best $hits@1000$ per dataset is \underline{underlined}. Best NDKL per dataset is \textbf{bold}. OOM = method ran out-of-memory.}
\begin{tabular}{llcccccc}
\toprule
\textbf{Method} &  & \textbf{Facebook} & \textbf{Credit} & \textbf{German} & \textbf{NBA} & \textbf{Pokec-n} & \textbf{Pokec-z} \\
\midrule

\multirow{2}{*}{UGE}
& NDKL
& 0.05 $\pm$ 0.00
& 0.80 $\pm$ 0.07
& 0.08 $\pm$ 0.03
& 0.07 $\pm$ 0.02
& 0.06 $\pm$ 0.00
& 0.06 $\pm$ 0.00 \\
& $hits@1000$
& 0.92 $\pm$ 0.00
& 0.51 $\pm$ 0.01
& 0.36 $\pm$ 0.01
& 0.58 $\pm$ 0.03
& 0.05 $\pm$ 0.01
& 0.04 $\pm$ 0.01 \\
\midrule

\multirow{2}{*}{EDITS GAE}
& NDKL
& 0.21 $\pm$ 0.07
& 0.04 $\pm$ 0.02
& 0.08 $\pm$ 0.02
& 0.08 $\pm$ 0.01
& OOM & OOM \\
& $hits@1000$
& 0.79 $\pm$ 0.21
& 0.56 $\pm$ 0.38
& 0.48 $\pm$ 0.29
& 0.36 $\pm$ 0.15
& OOM & OOM \\
\midrule

\multirow{2}{*}{GRAPHAIR}
& NDKL
& 0.13 $\pm$ 0.03
& 0.67 $\pm$ 0.22
& 0.07 $\pm$ 0.02
& 0.09 $\pm$ 0.01
& 0.09 $\pm$ 0.03
& 0.26 $\pm$ 0.25 \\
& $hits@1000$
& 0.85 $\pm$ 0.01
& 0.49 $\pm$ 0.05
& 0.39 $\pm$ 0.01
& 0.69 $\pm$ 0.02
& 0.11 $\pm$ 0.05
& 0.08 $\pm$ 0.12 \\
\midrule

\multirow{2}{*}{FairEGM}
& NDKL
& 0.09 $\pm$ 0.01
& 0.11 $\pm$ 0.00
& 0.05 $\pm$ 0.01
& 0.07 $\pm$ 0.01
& OOM & OOM \\
& $hits@1000$
& 0.95 $\pm$ 0.00
& 0.76 $\pm$ 0.01
& 0.29 $\pm$ 0.00
& 0.62 $\pm$ 0.02
& OOM & OOM \\
\midrule

\multirow{2}{*}{FairLP}
& NDKL
& 0.18 $\pm$ 0.00
& OOM
& 0.06 $\pm$ 0.00
& 0.20 $\pm$ 0.00
& OOM & OOM \\
& $hits@1000$
& 0.94 $\pm$ 0.00
& OOM
& 0.77 $\pm$ 0.00
& 0.83 $\pm$ 0.00
& OOM & OOM \\
\midrule

\multirow{2}{*}{FairWalk}
& NDKL
& 0.06 $\pm$ 0.01
& 0.06 $\pm$ 0.03
& 0.11 $\pm$ 0.02
& 0.06 $\pm$ 0.01
& 0.07 $\pm$ 0.01
& 0.07 $\pm$ 0.00 \\
& $hits@1000$
& 0.89 $\pm$ 0.01
& 0.92 $\pm$ 0.00
& 0.80 $\pm$ 0.00
& 0.54 $\pm$ 0.02
& 0.33 $\pm$ 0.00
& 0.32 $\pm$ 0.00 \\
\midrule

\multirow{2}{*}{FairAdj}
& NDKL
& 0.10 $\pm$ 0.05
& OOM
& 0.10 $\pm$ 0.01
& 0.11 $\pm$ 0.05
& OOM & OOM \\
& $hits@1000$
& 0.16 $\pm$ 0.01
& OOM
& 0.27 $\pm$ 0.01
& 0.51 $\pm$ 0.03
& OOM & OOM \\
\midrule

\multirow{2}{*}{DetConstSort GAE}
& NDKL
& 0.15 $\pm$ 0.00
& 0.06 $\pm$ 0.00
& 0.04 $\pm$ 0.00
& 0.09 $\pm$ 0.00
& 0.07 $\pm$ 0.00
& 0.23 $\pm$ 0.00 \\
& $hits@1000$
& \underline{1.00 $\pm$ 0.00}
& 0.00 $\pm$ 0.00
& \underline{1.00 $\pm$ 0.00}
& \underline{1.00 $\pm$ 0.00}
& \underline{1.00 $\pm$ 0.00}
& \underline{1.00 $\pm$ 0.00} \\
\midrule

\multirow{2}{*}{DELTR GAE}
& NDKL
& 0.10 $\pm$ 0.03
& 0.03 $\pm$ 0.00
& 0.09 $\pm$ 0.06
& 0.09 $\pm$ 0.02
& 0.23 $\pm$ 0.23
& 0.22 $\pm$ 0.20 \\
& $hits@1000$
& 0.62 $\pm$ 0.05
& 0.00 $\pm$ 0.00
& 0.12 $\pm$ 0.17
& 0.39 $\pm$ 0.26
& 0.01 $\pm$ 0.00
& 0.01 $\pm$ 0.01 \\
\midrule

\multirow{2}{*}{ThreeClass GAE}
& NDKL
& \textbf{0.04 $\pm$ 0.00}
& \textbf{0.01 $\pm$ 0.00}
& \textbf{0.03 $\pm$ 0.00}
& \textbf{0.03 $\pm$ 0.00}
& \textbf{0.02 $\pm$ 0.00}
& \textbf{0.03 $\pm$ 0.00} \\
& $hits@1000$
& \underline{1.00 $\pm$ 0.00}
& \underline{1.00 $\pm$ 0.00}
& \underline{1.00 $\pm$ 0.00}
& \underline{1.00 $\pm$ 0.00}
& \underline{1.00 $\pm$ 0.00}
& \underline{1.00 $\pm$ 0.00} \\
\bottomrule

\end{tabular}
\label{tab:hits_ndkl}
\end{table*}

\begin{table*}
\centering
\small
\setlength{\tabcolsep}{4pt}
\caption{Performance comparison (NDCG↑ and NDKL↓). Best NDCG per dataset is \underline{underlined}. Best NDKL per dataset is \textbf{bold}. OOM = method ran out-of-memory.}
\begin{tabular}{llcccccc}
\toprule
\textbf{Method} &  & \textbf{Facebook} & \textbf{Credit} & \textbf{German} & \textbf{NBA} & \textbf{Pokec-n} & \textbf{Pokec-z} \\
\midrule

\multirow{2}{*}{UGE}
& NDKL
& 0.05 $\pm$ 0.00
& 0.80 $\pm$ 0.07
& 0.08 $\pm$ 0.03
& 0.07 $\pm$ 0.02
& 0.06 $\pm$ 0.00
& 0.06 $\pm$ 0.00 \\
& NDCG
& \underline{0.99 $\pm$ 0.00}
& \underline{0.98 $\pm$ 0.00}
& 0.94 $\pm$ 0.00
& 0.92 $\pm$ 0.00
& 0.96 $\pm$ 0.00
& 0.96 $\pm$ 0.00 \\
\midrule

\multirow{2}{*}{EDITS GAE}
& NDKL
& 0.21 $\pm$ 0.07
& 0.04 $\pm$ 0.02
& 0.08 $\pm$ 0.02
& 0.08 $\pm$ 0.01
& OOM & OOM \\
& NDCG
& 0.97 $\pm$ 0.00
& 0.95 $\pm$ 0.00
& 0.91 $\pm$ 0.00
& 0.90 $\pm$ 0.00
& OOM & OOM \\
\midrule

\multirow{2}{*}{GRAPHAIR}
& NDKL
& 0.13 $\pm$ 0.03
& 0.67 $\pm$ 0.22
& 0.07 $\pm$ 0.02
& 0.09 $\pm$ 0.01
& 0.09 $\pm$ 0.03
& 0.26 $\pm$ 0.25 \\
& NDCG
& \underline{0.99 $\pm$ 0.00}
& \underline{0.99 $\pm$ 0.00}
& 0.94 $\pm$ 0.00
& 0.94 $\pm$ 0.00
& 0.97 $\pm$ 0.01
& 0.96 $\pm$ 0.03 \\
\midrule

\multirow{2}{*}{FairEGM}
& NDKL
& 0.09 $\pm$ 0.01
& 0.11 $\pm$ 0.00
& 0.05 $\pm$ 0.01
& 0.07 $\pm$ 0.01
& OOM & OOM \\
& NDCG
& \underline{0.99 $\pm$ 0.00}
& \underline{1.00 $\pm$ 0.00}
& \underline{0.93 $\pm$ 0.00}
& 0.92 $\pm$ 0.00
& OOM & OOM \\
\midrule

\multirow{2}{*}{FairLP}
& NDKL
& 0.18 $\pm$ 0.00
& OOM
& 0.06 $\pm$ 0.00
& 0.20 $\pm$ 0.00
& OOM & OOM \\
& NDCG
& \underline{0.99 $\pm$ 0.00}
& OOM
& \underline{0.98 $\pm$ 0.00}
& \underline{0.97 $\pm$ 0.00}
& OOM & OOM \\
\midrule

\multirow{2}{*}{FairWalk}
& NDKL
& 0.06 $\pm$ 0.01
& 0.06 $\pm$ 0.03
& 0.11 $\pm$ 0.02
& 0.06 $\pm$ 0.01
& 0.07 $\pm$ 0.01
& 0.07 $\pm$ 0.00 \\
& NDCG
& \underline{0.99 $\pm$ 0.00}
& \underline{1.00 $\pm$ 0.00}
& \underline{0.98 $\pm$ 0.00}
& 0.91 $\pm$ 0.00
& \underline{0.99 $\pm$ 0.00}
& \underline{0.99 $\pm$ 0.00} \\
\midrule

\multirow{2}{*}{FairAdj}
& NDKL
& 0.10 $\pm$ 0.05
& OOM
& 0.10 $\pm$ 0.01
& 0.11 $\pm$ 0.05
& OOM & OOM \\
& NDCG
& 0.91 $\pm$ 0.00
& OOM
& 0.91 $\pm$ 0.00
& 0.91 $\pm$ 0.00
& OOM & OOM \\
\midrule

\multirow{2}{*}{DetConstSort GAE}
& NDKL
& 0.15 $\pm$ 0.00
& 0.06 $\pm$ 0.00
& 0.04 $\pm$ 0.00
& 0.09 $\pm$ 0.00
& 0.07 $\pm$ 0.00
& 0.23 $\pm$ 0.00 \\
& NDCG
& 0.23 $\pm$ 0.00
& 0.00 $\pm$ 0.00
& 0.89 $\pm$ 0.00
& 0.77 $\pm$ 0.00
& 0.51 $\pm$ 0.00
& 0.28 $\pm$ 0.00 \\
\midrule

\multirow{2}{*}{DELTR GAE}
& NDKL
& 0.10 $\pm$ 0.03
& 0.03 $\pm$ 0.00
& 0.09 $\pm$ 0.06
& 0.09 $\pm$ 0.02
& 0.23 $\pm$ 0.23
& 0.22 $\pm$ 0.20 \\
& NDCG
& 0.97 $\pm$ 0.01
& 0.94 $\pm$ 0.00
& 0.91 $\pm$ 0.01
& 0.88 $\pm$ 0.06
& 0.93 $\pm$ 0.02
& 0.92 $\pm$ 0.02 \\
\midrule

\multirow{2}{*}{ThreeClass GAE}
& NDKL
& \textbf{0.04 $\pm$ 0.00}
& \textbf{0.01 $\pm$ 0.00}
& \textbf{0.03 $\pm$ 0.00}
& \textbf{0.03 $\pm$ 0.00}
& \textbf{0.02 $\pm$ 0.00}
& \textbf{0.03 $\pm$ 0.00} \\
& NDCG
& \underline{0.99 $\pm$ 0.00}
& \underline{1.00 $\pm$ 0.00}
& \underline{0.99 $\pm$ 0.00}
& \underline{0.97 $\pm$ 0.00}
& \underline{1.00 $\pm$ 0.00}
& \underline{1.00 $\pm$ 0.00} \\
\bottomrule

\end{tabular}
\label{tab:ndcg_ndkl}
\end{table*}

\begin{table*}
\centering
\small
\setlength{\tabcolsep}{4pt}
\caption{Fairness comparison using $\Delta_{DP}$↓ and Precision↑. Best (lowest) $\Delta_{DP}$ per dataset is \textbf{bold}. Best (highest) Precision per dataset is \underline{underlined}. OOM = method ran out-of-memory.}
\begin{tabular}{llcccccc}
\toprule
\textbf{Method} &  & \textbf{Facebook} & \textbf{Credit} & \textbf{German} & \textbf{NBA} & \textbf{Pokec-n} & \textbf{Pokec-z} \\
\midrule

\multirow{2}{*}{UGE}
& $\Delta_{DP}$
& \textbf{0.00 $\pm$ 0.00}
& 0.03 $\pm$ 0.00
& 0.01 $\pm$ 0.00
& \textbf{0.00 $\pm$ 0.00}
& 0.02 $\pm$ 0.00
& 0.02 $\pm$ 0.00 \\
& Precision
& \underline{0.97 $\pm$ 0.00}
& \underline{1.00 $\pm$ 0.00}
& 0.69 $\pm$ 0.01
& 0.58 $\pm$ 0.01
& 0.90 $\pm$ 0.04
& 0.91 $\pm$ 0.04 \\
\midrule

\multirow{2}{*}{EDITS GAE}
& $\Delta_{DP}$
& 0.01 $\pm$ 0.00
& 0.06 $\pm$ 0.04
& 0.01 $\pm$ 0.01
& \textbf{0.00 $\pm$ 0.00}
& OOM & OOM \\
& Precision
& 0.96 $\pm$ 0.00
& 0.36 $\pm$ 0.17
& 0.42 $\pm$ 0.20
& 0.49 $\pm$ 0.00
& OOM & OOM \\
\midrule

\multirow{2}{*}{GRAPHAIR}
& $\Delta_{DP}$
& \textbf{0.00 $\pm$ 0.00}
& \textbf{0.00 $\pm$ 0.00}
& \textbf{0.00 $\pm$ 0.00}
& \textbf{0.00 $\pm$ 0.00}
& \textbf{0.00 $\pm$ 0.00}
& 0.01 $\pm$ 0.01 \\
& Precision
& 0.96 $\pm$ 0.01
& \underline{1.00 $\pm$ 0.00}
& 0.73 $\pm$ 0.01
& 0.69 $\pm$ 0.01
& 0.97 $\pm$ 0.03
& \underline{1.00 $\pm$ 0.00} \\
\midrule

\multirow{2}{*}{FairEGM}
& $\Delta_{DP}$
& 0.01 $\pm$ 0.00
& 0.02 $\pm$ 0.00
& 0.01 $\pm$ 0.00
& 0.03 $\pm$ 0.00
& OOM & OOM \\
& Precision
& \underline{0.97 $\pm$ 0.00}
& \underline{1.00 $\pm$ 0.00}
& 0.62 $\pm$ 0.00
& 0.60 $\pm$ 0.01
& OOM & OOM \\
\midrule

\multirow{2}{*}{FairLP}
& $\Delta_{DP}$
& \textbf{0.00 $\pm$ 0.00}
& OOM
& 0.01 $\pm$ 0.00
& 0.01 $\pm$ 0.00
& OOM & OOM \\
& Precision
& 0.99 $\pm$ 0.00
& OOM
& 0.97 $\pm$ 0.00
& 0.86 $\pm$ 0.00
& OOM & OOM \\
\midrule

\multirow{2}{*}{FairWalk}
& $\Delta_{DP}$
& \textbf{0.00 $\pm$ 0.00}
& 0.01 $\pm$ 0.00
& 0.02 $\pm$ 0.00
& \textbf{0.00 $\pm$ 0.00}
& 0.03 $\pm$ 0.00
& 0.03 $\pm$ 0.00 \\
& Precision
& 0.96 $\pm$ 0.00
& \underline{1.00 $\pm$ 0.00}
& \underline{0.94 $\pm$ 0.00}
& 0.55 $\pm$ 0.01
& \underline{1.00 $\pm$ 0.00}
& \underline{1.00 $\pm$ 0.00} \\
\midrule

\multirow{2}{*}{FairAdj}
& $\Delta_{DP}$
& 0.15 $\pm$ 0.12
& OOM
& 0.52 $\pm$ 0.24
& 0.30 $\pm$ 0.23
& OOM & OOM \\
& Precision
& 0.42 $\pm$ 0.01
& OOM
& 0.54 $\pm$ 0.01
& 0.50 $\pm$ 0.01
& OOM & OOM \\
\midrule

\multirow{2}{*}{DetConstSort GAE}
& $\Delta_{DP}$
& \textbf{0.00 $\pm$ 0.00}
& \textbf{0.00 $\pm$ 0.00}
& \textbf{0.00 $\pm$ 0.00}
& 0.01 $\pm$ 0.00
& 0.01 $\pm$ 0.00
& 0.02 $\pm$ 0.00 \\
& Precision
& 0.00 $\pm$ 0.00
& 0.00 $\pm$ 0.00
& 0.55 $\pm$ 0.00
& 0.21 $\pm$ 0.00
& 0.07 $\pm$ 0.00
& 0.01 $\pm$ 0.00 \\
\midrule

\multirow{2}{*}{DELTR GAE}
& $\Delta_{DP}$
& \textbf{0.00 $\pm$ 0.00}
& 0.03 $\pm$ 0.02
& 0.03 $\pm$ 0.01
& 0.01 $\pm$ 0.02
& 0.05 $\pm$ 0.05
& 0.04 $\pm$ 0.03 \\
& Precision
& 0.91 $\pm$ 0.05
& 0.56 $\pm$ 0.29
& 0.31 $\pm$ 0.44
& 0.43 $\pm$ 0.24
& 0.65 $\pm$ 0.01
& 0.48 $\pm$ 0.28 \\
\midrule

\multirow{2}{*}{ThreeClass GAE}
& $\Delta_{DP}$
& \textbf{0.00 $\pm$ 0.00}
& \textbf{0.00 $\pm$ 0.00}
& 0.01 $\pm$ 0.00
& 0.07 $\pm$ 0.00
& \textbf{0.00 $\pm$ 0.00}
& \textbf{0.00 $\pm$ 0.00} \\
& Precision
& 0.95 $\pm$ 0.01
& \underline{1.00 $\pm$ 0.00}
& \underline{0.96 $\pm$ 0.00}
& \underline{0.80 $\pm$ 0.00}
& 0.98 $\pm$ 0.00
& 0.98 $\pm$ 0.00 \\
\bottomrule

\end{tabular}
\label{tab:parity_precision}
\end{table*}

\end{document}